\documentclass[11pt,a4paper]{article}
\ifdefined\pdfminorversion\pdfminorversion=7\fi

\usepackage[margin=0.82in]{geometry}

\usepackage[T1]{fontenc}

\usepackage[utf8]{inputenc}

\usepackage{lmodern,microtype}

\usepackage{amsmath,amssymb,amsthm,graphicx,booktabs,array,float,tabularx,enumitem}

\usepackage[font=small,labelfont=normalfont]{caption}

\usepackage[
    backend=biber,
    style=apa,
    sorting=nyt,
    natbib=true
]{biblatex}

\usepackage{xcolor}

\usepackage{algorithm}
\usepackage[noend]{algpseudocode}

\usepackage[colorlinks=true,linkcolor=blue!45!black,citecolor=blue!45!black,urlcolor=blue!45!black]{hyperref}

\hypersetup{
  pdftitle={Efficient Poincare Convolutions through Patch-Statistics Factorization},
  pdfauthor={Lukas Jochymek, Marketa Vasinkova, Petr Gajdos},
  pdfkeywords={hyperbolic neural networks; Poincare convolution;
    factorization; training efficiency; memory usage;
    convolution algorithms}
}

\setlist[itemize]{nosep,topsep=4pt,leftmargin=*}
\setlist[enumerate]{nosep,topsep=4pt,leftmargin=*}
\newcolumntype{Y}{>{\raggedright\arraybackslash}X}
\newcommand{\speed}[3]{#1\;[#2,\,#3]}
\newcommand{\bestspeed}[3]{\mathbf{#1}\;[#2,\,#3]}

\newtheorem{proposition}{Proposition}

\DeclareMathOperator{\asinh}{asinh}

\DeclareMathOperator{\atanh}{atanh}

\newcommand{\norm}[1]{\left\lVert#1\right\rVert}

\newcommand{\R}{\mathbb{R}}

\newcommand{\B}{\mathbb{B}}

\title{
  \vspace{-1.2em}
  Efficient Poincar\'e Convolutions through
  Patch-Statistics Factorization
}

\author{
  Luk\'a\v{s} Jochymek$^{a}$,
  Mark\'eta Va\v{s}inkov\'a$^{a,*}$,
  Petr Gajdo\v{s}$^{a}$\\[0.8em]
  \small $^{a}$Department of Computer Science,\\
  \small Faculty of Electrical Engineering and Computer Science,\\
  \small VSB--Technical University of Ostrava,\\
  \small 17. listopadu 2172/15, 708 00 Ostrava-Poruba, Czech Republic\\[0.8em]
  \small $^{*}$Corresponding author: Mark\'eta Va\v{s}inkov\'a\\
  \small E-mail: \texttt{marketa.vasinkova@vsb.cz}
}

\date{}

\usepackage[section]{placeins}
\begin{document}

\maketitle

\begin{abstract}
Published implementations of the Poincar\'e convolution in Hyperbolic Neural Networks++ (HNN++) map every overlapping image patch to the Poincar\'e ball before evaluating its score. We show that the clipped score depends on each patch only through two statistics, its squared norm and its inner products with the filters, which a native convolution and a single-channel box filter compute directly. The factorized layer therefore evaluates the specified score exactly, without constructing ball-valued patches, and its single-precision outputs and gradients agree with double-precision evaluations at least as closely as those of the explicit reference. On an RTX A6000 under two versions of JAX, complete training updates of a Poincar\'e U-Net and a Poincar\'e ResNet-20 are 1.5 to 1.8 times faster than the faster of two explicit implementations that map patches to the ball (1.2 to 1.4 times against an explicit implementation that skips this mapping); on a V100, the gain over the gather-based one persists as the ResNet-20 is widened eightfold. In 15 paired training runs per task, updates are 2.0 and 1.8 times faster than a reference that follows the published execution order, with no consistent change in final segmentation or classification quality, and peak memory falls by 38 to 55\% in a U-Net scaling study. Replacing the convolutions of the official PyTorch Hyperbolic U-Net makes its training step 1.4 times faster with half the memory. Evaluating an existing layer from its patch statistics thus makes Poincar\'e convolutional networks cheaper to train in the tested models and software stacks.
\end{abstract}


\section{Introduction}
Hyperbolic neural networks use negative curvature to represent hierarchical structure and have expanded from embedding models to complete vision architectures \citep{nickel2017,khrulkov2020,peng2022,mettes2024}, alongside hyperbolic kernel methods and graph convolutions \citep{fang2025,ye2025}. Their practical use also depends on the cost of the geometric operations surrounding ordinary tensor contractions. In a convolutional network, even modest overhead is repeated over many spatial locations and layers. Reducing this cost can make an existing model easier to train without requiring a different architecture or additional parameters.

The convolution in Hyperbolic Neural Networks++ (HNN++) provides a concrete opportunity \citep{shimizu2021}. A direct implementation concatenates the tangent features in each receptive field, maps the resulting patch to the Poincar\'e ball, and evaluates a hyperplane score. Overlapping receptive fields repeat much of this construction. Yet the score depends on the patch only through its squared norm and its inner products with the filters. The full ball-valued patch is therefore unnecessary for evaluating the score.

We exploit this structure by computing the directional products with a native convolution and the squared norms with a scalar energy map. The resulting factorization retains the source-radius cap, filter-norm floor, and output calculation of the reference layer. We call these two quantities the \emph{patch statistics}. The same radius and inner-product formulation also applies to HNN++ dense layers used in the complete models.

The contribution is an execution method for an existing Poincar\'e layer:
\begin{itemize}
\item An exact factorization of the clipped HNN++ convolution score into patch energy and directional products, computed by a native convolution and a scalar box filter (Section~\ref{sec:factorization}, Algorithm~\ref{alg:factorized}), together with the conditions under which the same evaluation applies to other layers, including Lorentz direct concatenation (Section~\ref{sec:discussion}).
\item A numerical evaluation that compares the factorized layer with an explicit reference in FP32 and FP64, tests FP32 gradients against FP64 evaluations at 40 trained states (Sections~\ref{sec:checks} and~\ref{sec:agreement}).
\item Measurements of operators, complete models, memory, and paired training cohorts on two GPUs (Sections~\ref{sec:operators}, \ref{sec:unet}, and~\ref{sec:training}).
\item Controls against gather-based and geometry-free explicit references under two JAX versions, a Euclidean network of the same topology, and the official HypLL layer in PyTorch (Sections~\ref{sec:attribution} and~\ref{sec:hypll}).
\end{itemize}

\section{Related work}
\citet{ganea2018} introduced Poincar\'e analogues of neural network operations through exponential and logarithmic maps. HNN++ developed fully connected and convolutional layers with a hyperplane interpretation \citep{shimizu2021}. Poincar\'e ResNet added residual architectures, midpoint batch normalization, and manual derivatives for geometric primitives \citep{spengler2023}. Hyperbolic U-Net extended this family to medical image segmentation and examined its initialization \citep{mishra2026}. These studies provide the layer definitions and architectural components used in our controlled models. Geoopt and HypLL provide software support for hyperbolic learning \citep{kochurov2020,spengler2023hypll}. These approaches motivate careful attention to execution cost, but they change the layer construction, as do Euclidean alternatives to standard convolution that redesign the operator to strengthen feature extraction at a similar parameter count \citep{jiang2025}; in contrast, we retain the layer and change only how it is evaluated.

Recent work has also addressed computational efficiency by developing alternative hyperbolic layers. Lorentz networks operate directly on the hyperboloid \citep{chen2022,bdeir2024}. \citet{bdeir2024robust} reparameterize the Lorentz linear map as a rotation followed by a boost, so that a native convolution performs the rotation and no explicit patch is built; this concerns the Lorentz model and its own parameterization, whereas we keep the Poincar\'e HNN++ layer and its parameters unchanged. The Intrinsic Lorentz Neural Network concatenates scaled patch features in the Lorentz model \citep{shi2026ilnn}, and Fast and Geometrically Grounded Lorentz Neural Networks combines a distance-to-hyperplane formulation with algebraic simplifications, Lorentzian activations, and parameter caching for inference \citep{klis2026}. Hyperbolic Busemann Neural Networks build efficient classification and fully connected layers from Busemann functions \citep{chen2026busemann}. These approaches reduce cost by changing the layer construction. Closest to our approach, hyperbolic image segmentation evaluates a Poincar\'e classifier at every pixel by expanding M\"obius addition into squared norms and inner products \citep{atigh2022}. We apply this principle to overlapping convolution patches, where it removes the explicit geometric patch tensor while retaining the specified clipped spatial HNN++ score. Clipping the feature norm before the exponential map is also used deliberately to improve hyperbolic classifiers \citep{guo2022clipped}; the radius cap considered here is instead the numerical safeguard of the reference layer, which we preserve unchanged.

\citet{mishne2023} analyze numerical limitations of hyperbolic representation learning. Recent work on graph networks with learned curvature also distinguishes task performance from geometric fidelity \citep{chen2026geometry}. We distinguish exact score preservation, numerical agreement, and training outcomes: an algebraic identity alone cannot establish agreement of implemented derivatives or subsequent optimization.

\section{Method}\label{sec:method}
\begin{figure}[H]

 \centering

 \includegraphics[width=\linewidth]{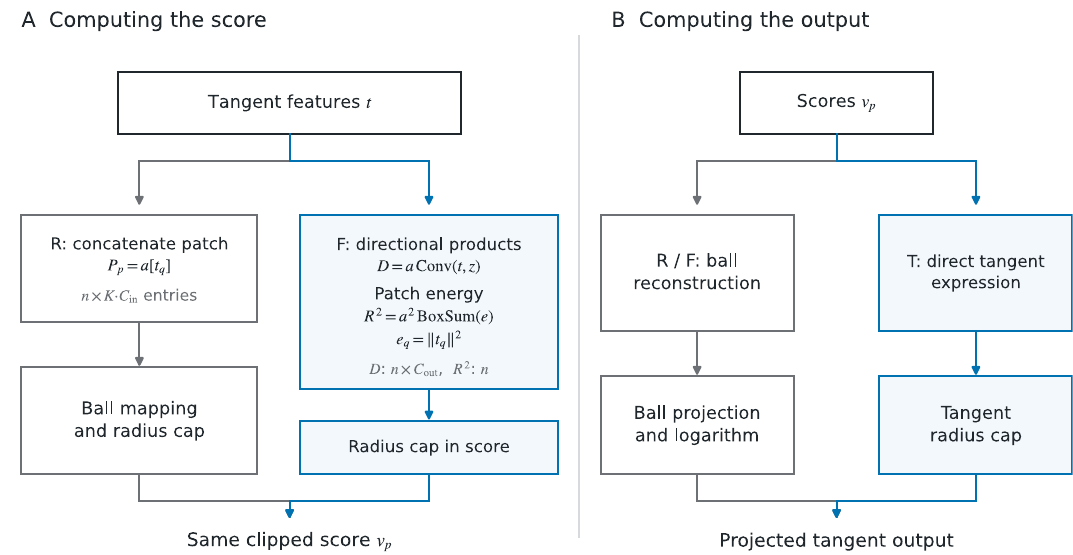}

 \caption{Compared execution paths. R materializes a beta-scaled tangent patch and maps it to the ball; F obtains its directional products and energy by convolution and scalar spatial reduction before evaluating the same clipped score. T additionally evaluates the projected tangent output directly and is used only as an ablation.}

 \label{fig:overview}

\end{figure}

We compare a reference implementation (R) that explicitly constructs geometric patches with the factorized implementation (F), which evaluates the same clipped score from patch energy and filter products. F retains the reference output calculation and is the primary method. A secondary ablation (T) also simplifies the output calculation. Figure~\ref{fig:overview} shows the execution paths. R is a standalone JAX implementation of the specified clipped HNN++ layer; the controlled model choices relative to upstream implementations are listed in Appendix~\ref{app:controlled}. R follows the execution order of public implementations. The official HNN++ convolution concatenates shifted, beta-scaled tangent features and maps the concatenation to the ball \citep{hnnppcode}. The Poincar\'e ResNet convolution and HypLL's \texttt{HConvolution2d}, which Hyperbolic U-Net uses, unfold the complete beta-scaled tangent patch and map it to the ball before evaluating the fully connected score \citep{poincareresnetcode,hypllcode}. None of these pinned revisions evaluates the score from patch statistics.

R builds the patch with JAX's one-hot patch convolution followed by a transpose. Three benchmark controls complete the comparison. Rg builds the bitwise-identical patch by gathering shifted slices, as the official HNN++ code does, and is otherwise R. Mg keeps Rg's patch but computes its energy and directional products directly from the patch and evaluates F's score; Rg versus Mg therefore isolates the pointwise ball geometry, and Mg versus F isolates native convolution of the patch statistics. E is a Euclidean network of the same topology with parameter-shape-matched convolution and dense layers, current-batch normalization over the same axes, and ordinary skip concatenation and residual addition. E serves only as a cost reference; its task quality is not evaluated.

\subsection{Notation and reference layer}

All norms and inner products are Euclidean unless stated otherwise, and tangent coordinates are based at the origin. Let $c>0$ be the curvature magnitude, so the sectional curvature is $-c$. On $\B_c^d=\{x\in\R^d:c\norm{x}^2<1\}$, use

\begin{equation}
 \exp_0^c(t)=\frac{\tanh(\sqrt c\norm{t})}{\sqrt c\norm{t}}t,
 \qquad
 \log_0^c(x)=\frac{\atanh(\sqrt c\norm{x})}{\sqrt c\norm{x}}x,
 \label{eq:maps}
\end{equation}

with analytic extensions at zero. The dimension of the ball is $d$, $x$ is a ball point, and $t$ is an origin-tangent vector. Let $B$ denote batch size, $H_{\mathrm{in}},W_{\mathrm{in}}$ the input spatial dimensions, $H_{\mathrm{out}},W_{\mathrm{out}}$ the output dimensions, and $C_{\mathrm{in}},C_{\mathrm{out}}$ the channel counts. We suppress the batch index below. Write $t_q=\log_0^c(x_q)\in\R^{C_{\mathrm{in}}}$ for the feature at input location $q$. For output location $p$, let $\mathcal P_p$ be the ordered receptive-field index set under the chosen stride and padding. A $k\times k$ kernel has $K=k^2$ offsets. With $\mathrm{B}(\cdot,\cdot)$ denoting Euler's beta function, set $\beta_d=\mathrm{B}(d/2,1/2)$ and $a=\beta_{KC_{\mathrm{in}}}/\beta_{C_{\mathrm{in}}}$, a layer constant close to $1/\sqrt K$ (0.31--0.33 for $k=3$). The beta-scaled patch is

\begin{equation}
 P_p=a[\,t_q\,]_{q\in\mathcal P_p}.
 \label{eq:patch}
\end{equation}

Here, brackets denote concatenation in kernel-offset order, so $P_p\in\R^{KC_{\mathrm{in}}}$. The reference maps $P_p$ to the ball before evaluating the HNN++ hyperplane score. For output channel $j\in\{1,\ldots,C_{\mathrm{out}}\}$, let $z_j\in\R^{KC_{\mathrm{in}}}$ be the concatenated filter and $r_j\in\R$ its scalar offset. The block $z_{\delta j}\in\R^{C_{\mathrm{in}}}$ corresponds to relative kernel offset $\delta$; it is shared across spatial positions. We write $\delta(p,q)$ for the offset that associates input location $q$ with output location $p$.

\subsection{Factorizing the patch computation}\label{sec:factorization}

The score uses the tangent patch only through

\begin{equation}
 R_p^2=\norm{P_p}^2=a^2\sum_{q\in\mathcal P_p}\norm{t_q}^2,
 \qquad
 D_{pj}=z_j^\top P_p=a\sum_{q\in\mathcal P_p}z_{\delta(p,q)j}^\top t_q.
 \label{eq:factorization}
\end{equation}

Here $R_p=\norm{P_p}$ is the nonnegative patch radius, $D_{pj}$ is its directional product with filter $j$, and $R_p^2$ is shared across output channels. The subscripted radius $R_p$ is distinct from the reference-path label R.

We compute $D$ with a native convolution and $R^2$ by convolving the channel-summed squared input with an all-ones scalar kernel. Both use the stride, padding, and dilation of the layer. Zero padding of tangent features corresponds to padding with the ball origin, as in the reference, and the filter blocks follow the cross-correlation offset order of the native convolution. The implementation evaluates the \emph{clipped} score. Let

\begin{gather}
 m=1-\epsilon,\qquad \bar g_j=\max(\norm{z_j},10^{-15}),
 \qquad \rho_p=\tanh(\sqrt c R_p),\qquad
 \bar\rho_p=\min(\rho_p,m),\label{eq:scoreclip}\\
 \eta_p=
 \begin{cases}
  \bar\rho_p/R_p,&R_p>0,\\
  \sqrt c,&R_p=0,
 \end{cases}
 \qquad \bar\lambda_p=\frac{2}{1-\bar\rho_p^2}.\nonumber
\end{gather}

The dimensionless radius $\rho_p$ is capped at $m$, $\eta_p$ converts the tangent directional product to its capped ball counterpart, and $\bar\lambda_p$ is the conformal factor at that ball point. The positive floor in $\bar g_j$ keeps the filter normalization defined. In particular, the projected patch is
\[
 x_p=\frac{\bar\rho_p}{\sqrt c R_p}P_p,\qquad
 \sqrt c\left\langle x_p,\frac{z_j}{\bar g_j}\right\rangle
 =\eta_p\frac{D_{pj}}{\bar g_j},
\]
with the zero-radius case defined by continuity. Substitution into the HNN++ score gives the expression shared by all three paths:

\begin{align}
 v_{pj}&=\frac{2\bar g_j}{\sqrt c}\asinh(A_{pj}),\label{eq:score}\\
 A_{pj}&=\bar\lambda_p\eta_p\frac{D_{pj}}{\bar g_j}
          \cosh(2\sqrt c r_j)
       -(\bar\lambda_p-1)\sinh(2\sqrt c r_j).\nonumber
\end{align}

The margin $\epsilon$ depends on precision: the specified implementation uses $\epsilon=0.004$ in FP32 and $10^{-5}$ in FP64; these and the $10^{-15}$ filter-norm floor define the evaluated layer. When clipping is inactive, $\bar\rho_p=\rho_p$ and the two coefficients reduce to $\sinh(2\sqrt c R_p)/R_p$ and $\cosh(2\sqrt c R_p)$, respectively. Thus Eq.~\eqref{eq:score} includes both the source-radius cap and the filter-norm floor as used by the reference implementation.

\begin{proposition}[Patch factorization]

For the patch in Eq.~\eqref{eq:patch}, Eq.~\eqref{eq:factorization} gives exactly the squared norm and directional products of the concatenated patch in real arithmetic. Substitution into Eq.~\eqref{eq:score} therefore preserves the specified clipped score wherever the reference expression is defined, and its derivatives with respect to inputs, filters, offsets, and curvature wherever that expression is differentiable. Floating-point evaluations of the score and its derivatives can still differ (Section~\ref{sec:agreement}).

\end{proposition}

\begin{proof}

The squared norm of a concatenation is the sum of its block squared norms, and its inner product with a concatenated filter is the sum of block inner products. Scaling every patch block by $a$ contributes $a^2$ and $a$, respectively.
\end{proof}

The identity is elementary: the origin exponential map rescales the patch radially, so the score depends on it only through its norm and filter products. Its value lies in evaluating these quantities without constructing the ball-valued patch tensor. In the final checkpoints of the V100 R and F training runs, evaluated on the official test sets, the radius cap is active for at most $3\times10^{-8}$ of all patch evaluations and at most $2\times10^{-5}$ of the evaluations in any single layer and checkpoint (Supplementary Table~S13), so the trained layers operate in the unclipped regime.

Algorithm~\ref{alg:factorized} summarizes the factorized layer. The reference instead forms the patch tensor with $nKC_{\mathrm{in}}$ entries, where $n=BH_{\mathrm{out}}W_{\mathrm{out}}$, and maps it to the ball before evaluating the same score.

\begin{algorithm}[htbp]
\caption{Factorized HNN++ Poincar\'e convolution (F).}
\label{alg:factorized}
\begin{algorithmic}[1]
\Require tangent features $t$ of shape $B\times H_{\mathrm{in}}\times W_{\mathrm{in}}\times C_{\mathrm{in}}$; filters $z$; offsets $r$; curvature $c$
\State $e_q\gets\norm{t_q}^2$ \Comment{energy map, $B\times H_{\mathrm{in}}\times W_{\mathrm{in}}$}
\State $R^2\gets a^2\,\mathrm{BoxSum}_{k\times k}(e)$ \Comment{$B\times H_{\mathrm{out}}\times W_{\mathrm{out}}$}
\State $D\gets a\,\mathrm{Conv}(t,z)$ \Comment{native convolution, $B\times H_{\mathrm{out}}\times W_{\mathrm{out}}\times C_{\mathrm{out}}$}
\State $\bar g_j\gets\max(\norm{z_j},10^{-15})$
\State $\bar\rho\gets\min(\tanh(\sqrt c\,R),m)$, \ $\eta\gets\bar\rho/R$ ($\sqrt c$ if $R=0$) \Comment{radius cap, Eq.~\eqref{eq:scoreclip}}
\State $v\gets$ Eq.~\eqref{eq:score} \Comment{scores, $B\times H_{\mathrm{out}}\times W_{\mathrm{out}}\times C_{\mathrm{out}}$}
\State \Return reference output map of $v$: reconstruction, projection, $\log_0^c$
\end{algorithmic}
\end{algorithm}

For an HNN++ dense layer there is no spatial patch. R explicitly maps its tangent input $t$ to the ball, whereas F and T pass $\norm{t}^2$ and the filter products $z_j^\top t$ to the same clipped-score evaluation. This dense transformation is part of the complete models: it is used in the ResNet classifier and in the fully connected step of transpose decoding, which is followed by folding. F retains the reference output calculation and all learned parameters.

Let $n=B H_{\mathrm{out}}W_{\mathrm{out}}$. R and F have the same leading dot-product complexity, $O(nK C_{\mathrm{in}}C_{\mathrm{out}})$; factorization does not lower the asymptotic convolution order. R also materializes $nK C_{\mathrm{in}}$ patch elements and applies pointwise geometry to them. F removes this explicit ball-patch construction and uses a directional-score tensor with $nC_{\mathrm{out}}$ entries; its scalar energy path costs $O(BH_{\mathrm{in}}W_{\mathrm{in}}C_{\mathrm{in}}+nK)$ through a channel reduction followed by a one-channel spatial convolution. A native convolution backend may still allocate workspaces. We therefore measure memory rather than inferring it from these element counts.

\subsection{Output calculation and implementation}
R and F reconstruct the ball-valued output, project it to the prescribed radius, and apply the logarithmic map. T evaluates the corresponding tangent expression directly with the matching radial cap. Its derivation and implementation are given in Appendix~\ref{app:direct}; this ablation separates the effect of patch factorization from the additional output simplification.

All paths retain the same projection margins, parameters, normalization, and initialization. Limiting expressions at zero, Taylor approximations, and derivative conventions at clipping boundaries are specified in Appendix~\ref{app:implementation}.

\section{Experimental design}

\subsection{Agreement checks}\label{sec:checks}
Experiments use JAX 0.4.25 \citep{jax2018}. Implementation checks compare outputs and reverse-mode gradients with respect to inputs, weights, offsets, and curvature in FP32 and FP64. For a reference tensor $y_{\mathrm{ref}}$ and candidate tensor $y_{\mathrm{cand}}$, the prespecified tensor-level gate is
\begin{equation}
 \norm{y_{\mathrm{cand}}-y_{\mathrm{ref}}}_{\infty}
 \leq \epsilon_{\mathrm{abs}}+\epsilon_{\mathrm{rel}}\norm{y_{\mathrm{ref}}}_2.
 \label{eq:tolerance}
\end{equation}
Both tolerances are $2\times10^{-5}$ in FP32 and $10^{-10}$ in FP64. The norms cover all entries of the compared tensor, so the allowance on the largest error grows with $\norm{y_{\mathrm{ref}}}_2$ and therefore with tensor size. We therefore report relative L2 errors $\norm{y_{\mathrm{cand}}-y_{\mathrm{ref}}}_2/\norm{y_{\mathrm{ref}}}_2$ as the primary summary and the gate as a secondary count. A separate replay of the structured suite also records the elementwise relative error $|y_{\mathrm{cand}}-y_{\mathrm{ref}}|/\max(|y_{\mathrm{ref}}|,10^{-12})$ for every compared tensor.

The structured tests use ten realizations per precision of an eight-layer convolution--ReLU stack and a small U-Net of depth two. Outputs are scalarized with seeded Gaussian cotangents before differentiation. Separate analytic and stress tests examine zero inputs, clipping boundaries, and curvature-related cancellation; these are distinguished from the smooth agreement tests. Curvatures from $0.001$ to $10$ occur across the validation suites, but are not crossed with every topology. These checks assess agreement with the reference under the specified tensor-level tolerance.

\subsection{Execution measurements}\label{sec:timing}
Operator, block, complete U-Net, memory, and scaling measurements use one NVIDIA Tesla V100 with 32~GiB of physical memory and CUDA 11.8. Timed computation uses FP32, the highest matrix-multiplication precision, and disabled JAX preallocation. The recorded allocator limit of 23.81~GiB is JAX's default fraction of 0.75 of the device memory available to CUDA, which also caps the allocator when preallocation is disabled. Convolution workloads S1--S3 use batch 8, $3\times3$ kernels, stride 1, and SAME padding. The convolution--normalization--ReLU block uses batch 2, spatial size $32^2$, and $32\to32$ channels. The dense stack has four layers of width 256 and batch size 256.

Each workload uses five paired sessions in fresh processes, with randomized variant order, 50 warmup calls, and 200 synchronized measurements per variant. Compilation precedes timing; inputs and parameters remain on the device. Every returned output or state leaf is synchronized. Forward and gradient measurements reuse fixed state. Gradient calls differentiate the mean squared output with respect to the parameter tree, input, and curvature, without an optimizer update. Complete U-Net updates instead advance the parameters and both Adam moments on a repeated synthetic batch. All paths start from identical newly initialized states within each session and then evolve independently.

Each session contributes its median latency. Speedups are geometric means of paired reference-to-candidate ratios, with unadjusted 95\% Student-$t$ intervals on the log ratios. Latency excludes initialization, compilation, warmup, and host transfer. Peak memory is the median of the five process-lifetime peak live allocations and includes initialization, compilation, and warmup. It therefore measures a broader boundary than latency. A separate 20-call trace describes device activity.

The U-Net scaling study varies batch size over $\{1,2,4,8,16\}$ at $256\times256$, and resolution over $\{128,256,512\}$ at batch 8, measuring the shared configuration once. It uses the same timing protocol and retains every planned session outcome. Out-of-memory outcomes require an explicit allocator diagnostic; paired speedups require completion by both paths. A repeat of three scaling workloads raises the allocator fraction to 0.95 under the same protocol. Software versions and a subsequent rerun on the same V100 are documented in Appendix~\ref{app:environment}.

The gather-based controls Rg and Mg and the Euclidean network E are measured with the same protocol on an RTX A6000 with an allocator fraction of 0.95, first with JAX 0.4.25 and then with JAX 0.11.2. R, Rg, Mg, and F share paired sessions; R, F, and E share separate paired sessions. The workloads are the batch-8 U-Net forward call and update and the ResNet-20 update at batch 128, the configurations used in training. On the V100, a width sweep of the ResNet-20 update compares R, Rg, and F; R and F are also compared at twice the U-Net width; and E is measured at the workloads of Table~\ref{tab:unettime} and Figure~\ref{fig:scaling}. These controls, the PyTorch comparison (Section~\ref{sec:hypll}), and the FP64-oracle analyses were added after the primary results were known. Device assignments and session exclusions are listed in Appendix~\ref{app:environment}.

\subsection{Training and statistical analysis}
We train a Poincar\'e U-Net on ISIC16 segmentation and a compact Poincar\'e ResNet-20 on CIFAR-100 classification \citep{gutman2016,krizhevsky2009}. The models follow the architectural families of U-Net, Hyperbolic U-Net, and Poincar\'e ResNet \citep{ronneberger2015,mishra2026,he2016,spengler2023}; the controlled settings are summarized in Table~\ref{tab:trainingdesign}. F factorizes both spatial scores and the dense scores in the U-Net decoder and ResNet classifier. Neither architecture nor parameter count changes between paths. Both models are compact controlled testbeds rather than accuracy benchmarks. A published Poincar\'e ResNet-20 with the same widths reaches 54.1\% CIFAR-100 top-1 accuracy after 500 epochs, and its Euclidean counterpart reaches 54.3\% \citep[Table~1]{spengler2025adversarial}; a Poincar\'e ResNet-18 reaches 76.6\% \citep[Table~1]{bdeir2024}.

\begin{table}[htbp]
\centering\small
\caption{Training settings shared within each task. Both models use midpoint normalization with current-batch statistics and no running state. Detailed recipes are given in Appendix~\ref{app:protocol}.}
\label{tab:trainingdesign}
\setlength{\tabcolsep}{5pt}
\begin{tabularx}{\linewidth}{@{}p{0.18\linewidth}YY@{}}
\toprule
\textbf{Setting} & \textbf{ISIC16: Poincar\'e U-Net} & \textbf{CIFAR-100: Poincar\'e ResNet-20}\\
\midrule
Data split & 810 training / 90 development / 379 test & 45,000 training / 5,000 development / 10,000 test\\
\addlinespace
Architecture & Depth 4; widths $(8,16,32,64,128)$ & Three residual groups; widths $(8,16,32)$\\
Parameters & 486,941; 19 spatial convolutions & 71,727; 21 spatial convolutions\\
Curvature & Learned, initialized at $c=0.1$ & Fixed at $c=0.1$\\
\addlinespace
Training budget & Batch 8; 50 epochs; 5,100 updates & Batch 128; 100 epochs; 35,200 updates\\
Loss & Dice and focal loss & Cross entropy\\
Optimizer & Adam, learning rate $10^{-3}$, coupled L2 $10^{-4}$ & Adam, learning rate $10^{-3}$, coupled L2 $10^{-4}$\\
Gradient clipping & Global norm 1, before L2 & None\\
\addlinespace
Primary test metric & Mean per-image foreground Dice & Top-1 accuracy\\
Evaluation batch & 8 & 128\\
\bottomrule
\end{tabularx}
\end{table}

The V100 cohorts compare R, F, and T using five seeds per task. The A6000 replications compare R and F using 15 seeds per task, specified before endpoint analysis. Within a seed, all paths share initialization and data order. F was designated the primary method after the complete-U-Net benchmark and before any training run. All 60 A6000 runs completed before the final endpoints were aggregated. The cohorts are analyzed separately because the platform, seed set, and protocol version differ.

The primary endpoint uses the checkpoint at the fixed final update. A secondary checkpoint is selected by development performance after each epoch; ties retain the earlier checkpoint. Training-update time sums synchronized optimizer calls, excluding compilation and evaluation. The separate elapsed worker time includes data materialization, initialization, compilation, training, development evaluation, checkpointing, final and selected-checkpoint test evaluation, and prediction writing. It excludes launcher checks, subprocess startup, and the final GPU snapshot.

Training speedups use paired seed ratios and the same log-$t$ interval as the execution benchmarks. Quality differences are paired candidate-minus-reference seed differences, with arithmetic means and approximate paired-$t$ 95\% intervals. These intervals describe variation across training seeds. No multiplicity correction or equivalence or noninferiority margin was prespecified. Appendix~\ref{app:secondary} summarizes sensitivity analyses and post hoc equivalence tests; Supplementary Tables~S4--S7 report individual seeds and extended outcomes. Sensitivity analyses exclude three ISIC16 and ten CIFAR-100 test inputs that are byte-identical to training inputs and reevaluate the remaining test sets with their resulting batch composition, since both models use current-batch normalization. The ISIC16 split does not establish patient or lesion independence.

\section{Results}

\FloatBarrier
\subsection{Agreement with the reference}\label{sec:agreement}
Table~\ref{tab:numerics} summarizes the structured comparisons. The median relative L2 error of F and T against R is at the level of floating-point rounding, about $10^{-6}$ in FP32 and $10^{-15}$ in FP64, for outputs and for gradients. Under the prespecified gate, F passes all 1,360 comparisons in the accepted run and 1,328 in a full replay, and T passes 1,327 and 1,360. Apart from one T curvature gradient that exceeds the gate by 8\%, every miss comes from one small-U-Net FP32 realization in which the exact value of one ReLU input exceeds zero by only four FP32 units in the last place at the tensor's scale. FP32 rounding decides the sign of this input, and the ReLU derivative then passes or blocks the gradient at that element, which changes several gradient tensors by up to 7.8\% relative L2. The accepted run and the replay were compiled separately from different code revisions, which can change the FP32 accumulation order; we did not isolate which operation changed. An FP64 oracle shows that R itself took the wrong side in the accepted run, so these misses reflect a rounding-determined ReLU decision rather than a failure of the real-arithmetic identity. Against this oracle, computed on CPU, the FP32 errors of F and R are comparable, with median relative L2 errors of about $10^{-6}$ for outputs and gradients at curvatures from 0.06 to 0.13 (Supplementary Table~S11). Stress diagnostics expose FP32 curvature-gradient cancellation in both R and F and sensitivity to branch selection at clipping boundaries, so the structured comparisons do not establish accurate derivatives in every regime.

Gradients at trained states were evaluated at 40 checkpoints of the V100 training runs: ResNet-20 states produced by R, F, and T at mid-training and at the end for five seeds, and U-Net states produced by R and F at the end. The reference is F evaluated in FP64; at an inspected ResNet-20 state it agrees with R evaluated in FP64 on the CPU to about $10^{-8}$ and with central finite differences of the loss. Against this reference, the median relative L2 error of the FP32 gradients of F and T is about $3\times10^{-5}$ for the U-Net and $3$--$9\times10^{-4}$ for ResNet-20 (Supplementary Table~S15). R reaches the same level for every tensor except the weight gradient of the ResNet-20 stem convolution, which the GPU executable compiled for R in our pinned software stack evaluates incorrectly; computing the weight normalization before the contraction removes this deviation, and forward values and the U-Net are unaffected (Supplementary Tables~S14--S16).

Elementwise relative errors (Supplementary Table~S9) are at most $1.4\times10^{-7}$ in FP64, where the absolute error is below $3\times10^{-13}$ in every tensor. In FP32 they are typically about $10^{-6}$; the largest values occur at gradient entries whose reference value is close to zero, where the absolute error is below $10^{-9}$.

\begin{table}[htbp]
\centering\small
\caption{Structured agreement of F and T with R in the accepted run and in a full replay. Each path has 1,360 compared tensors, 680 per precision: 20 outputs and 660 input, parameter, and curvature gradients. Passed counts use the tensor-level criterion of Eq.~\eqref{eq:tolerance} and cover both precisions. Error columns give the median and maximum across tensors of the relative L2 error $\norm{y_{\mathrm{cand}}-y_{\mathrm{ref}}}_2/\norm{y_{\mathrm{ref}}}_2$. Apart from one T curvature gradient in the accepted run, every miss comes from one small-U-Net FP32 realization whose outcome is set by the rounding of a single ReLU input (Section~\ref{sec:agreement}); this realization also gives the largest FP32 gradient error of T in the accepted run and of F in the replay. The other FP32 gradient maxima are convolution-stack curvature gradients with reference L2 norms below $3\times10^{-7}$.}
\label{tab:numerics}
\setlength{\tabcolsep}{4pt}
\begin{tabular}{@{}llrlcccc@{}}
\toprule
 & & & & \multicolumn{2}{c}{\textbf{Outputs}} & \multicolumn{2}{c}{\textbf{Gradients}}\\
\cmidrule(lr){5-6}\cmidrule(l){7-8}
\textbf{Path} & \textbf{Run} & \textbf{Passed} & \textbf{Precision} & \textbf{Median} & \textbf{Max} & \textbf{Median} & \textbf{Max}\\
\midrule
F & Accepted & 1,360 / 1,360 & FP32 & $1.1\times10^{-6}$ & $1.8\times10^{-6}$ & $8.2\times10^{-7}$ & $1.1\times10^{-2}$\\
 &  &  & FP64 & $1.8\times10^{-15}$ & $2.8\times10^{-15}$ & $1.4\times10^{-15}$ & $1.6\times10^{-11}$\\
 & Replay & 1,328 / 1,360 & FP32 & $1.1\times10^{-6}$ & $1.8\times10^{-6}$ & $8.5\times10^{-7}$ & $6.5\times10^{-2}$\\
 &  &  & FP64 & $1.8\times10^{-15}$ & $3.1\times10^{-15}$ & $1.4\times10^{-15}$ & $1.6\times10^{-11}$\\
\addlinespace
T & Accepted & 1,327 / 1,360 & FP32 & $1.3\times10^{-6}$ & $2.0\times10^{-6}$ & $9.3\times10^{-7}$ & $7.8\times10^{-2}$\\
 &  &  & FP64 & $1.9\times10^{-15}$ & $3.1\times10^{-15}$ & $1.5\times10^{-15}$ & $7.3\times10^{-12}$\\
 & Replay & 1,360 / 1,360 & FP32 & $1.3\times10^{-6}$ & $2.2\times10^{-6}$ & $9.2\times10^{-7}$ & $1.9\times10^{-2}$\\
 &  &  & FP64 & $2.0\times10^{-15}$ & $3.2\times10^{-15}$ & $1.5\times10^{-15}$ & $7.3\times10^{-12}$\\
\bottomrule
\end{tabular}
\end{table}

\FloatBarrier
\subsection{Operator and block runtime}\label{sec:operators}
Factorization accelerates all three spatial convolutions, with the largest gains at high spatial resolution (Table~\ref{tab:operators}). Improvements are smaller for the normalization block and dense stack, where more of the execution is unaffected. Both dense-stack intervals include one, so this isolated benchmark gives limited evidence of a dense-layer speedup. The direct tangent output (T) is compared with F in Appendix~\ref{app:secondary}; full results are in Supplementary Tables~S1--S3.

\begin{table}[htbp]
\centering\small
\caption{Operator and block speedups, measured as reference time divided by factorized time. Brackets show paired 95\% intervals. Bold estimates have intervals entirely above one. Gradient calls include reverse-mode differentiation but no optimizer update.}
\label{tab:operators}
\begin{tabular}{@{}lcc@{}}
\toprule
\textbf{Workload} & \textbf{Forward} & \textbf{Forward + gradient}\\
\midrule
S1: $16\times16$, $256\to256$ & $\bestspeed{1.83}{1.74}{1.93}$ & $\bestspeed{2.07}{1.94}{2.21}$\\
S2: $256\times256$, $32\to32$ & $\bestspeed{9.58}{9.48}{9.68}$ & $\bestspeed{3.38}{3.31}{3.46}$\\
S3: $256\times256$, $96\to32$ & $\bestspeed{25.81}{25.03}{26.62}$ & $\bestspeed{5.67}{5.58}{5.75}$\\
\addlinespace
Conv--BN--ReLU & $\bestspeed{1.17}{1.02}{1.33}$ & $\speed{1.05}{0.98}{1.13}$\\
Four-layer dense stack & $\speed{1.09}{0.94}{1.27}$ & $\speed{1.10}{0.998}{1.22}$\\
\bottomrule
\end{tabular}
\end{table}

\FloatBarrier
\subsection{Complete U-Net runtime and memory}\label{sec:unet}
The benefit extends to complete model execution. For a batch of eight, F nearly halves U-Net update time and reduces peak live allocation by 54\% (Table~\ref{tab:unettime}). The memory effect depends on the workload: at batch one, forward allocation is higher for F. This difference comes from convolution autotuning during compilation, which the peak includes; with autotuning disabled, F's batch-1 peak falls below R's.\footnote{Batch-1 forward peak allocation with convolution autotuning disabled, median of five sessions: 0.019~GiB for F and 0.128~GiB for R.} The Euclidean U-Net of the same topology completes the batch-8 update in 23.5~ms with 0.62~GiB, so F's update takes 2.7 times as long and R's 5.3 times.

\begin{table}[htbp]
\centering\small
\caption{Complete U-Net execution on V100. Bold speedups have 95\% intervals that exclude one. E is a Euclidean U-Net of the same topology, measured alone on a second, identical V100 as a cost reference; forward peaks are dominated by compilation workspace, which F and E share. Latency excludes initialization, compilation, and warmup; peak allocation includes them. Paired speedups can differ from the quotient of the displayed median latencies.}
\label{tab:unettime}
\setlength{\tabcolsep}{5pt}
\begin{tabular}{@{}llrcr@{}}
\toprule
\textbf{Workload} & \textbf{Path} & \textbf{Time (ms)} & \textbf{Speedup [95\% interval]} & \textbf{Peak (GiB)}\\
\midrule
Batch 1, forward & R & 9.83 & -- & 0.237\\
 & F & 5.14 & $\bestspeed{1.86}{1.79}{1.94}$ & 0.395\\
 & E & 1.24 & -- & 0.395\\
\addlinespace
Batch 8, forward & R & 48.44 & -- & 0.885\\
 & F & 17.86 & $\bestspeed{2.69}{2.61}{2.77}$ & 0.619\\
 & E & 3.59 & -- & 0.619\\
\addlinespace
Batch 8, update & R & 124.06 & -- & 5.955\\
 & F & 64.32 & $\bestspeed{1.92}{1.75}{2.11}$ & 2.721\\
 & E & 23.50 & -- & 0.623\\
\bottomrule
\end{tabular}
\end{table}

\begin{figure}[htbp]
\centering
\includegraphics[width=\linewidth]{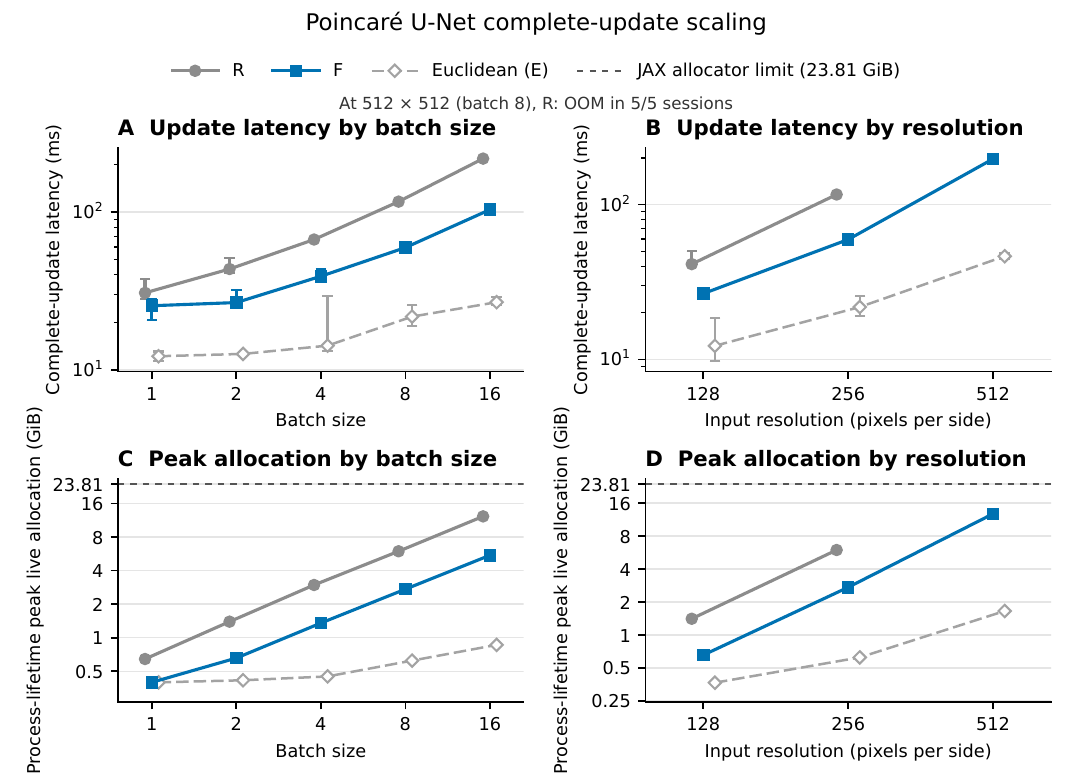}
\caption{U-Net update scaling on V100. Batch size varies at $256\times256$ in A and C; resolution varies at batch 8 in B and D. Points show session medians with latency bars spanning the session minimum and maximum; R and F use five sessions, and E uses the sessions retained under the driver-stall rule of Appendix~\ref{app:environment} (two at batch 1 and 2, five elsewhere). E points come from separate sessions at these workloads, so at batch 8 and $256\times256$ the E point differs from the E entry of Table~\ref{tab:unettime}. Peak allocation covers the process lifetime. At $512\times512$, R ran out of memory in all five sessions, and no R value is plotted. The gray dashed line is a Euclidean U-Net of the same topology (E), measured alone. The black dashed line is the recorded JAX allocator limit, 75\% of the device memory available to CUDA, below the device's 32~GiB physical memory.}
\label{fig:scaling}
\end{figure}

The scaling study tests whether these improvements persist as batch size or resolution increases (Figure~\ref{fig:scaling}). F is faster in all six configurations completed by both paths, with paired speedups from $1.29\times$ to $2.10\times$ and 38.3--55.2\% lower peak allocation. At batch 8 and $256\times256$, the scaling study reproduces the earlier benchmark closely: the speedup is $1.935$ [1.907, 1.964], and peak allocations are again 5.955~GiB for R and 2.721~GiB for F. Across four separate V100 measurements of this update, the speedup ranges from 1.75 to 1.94, so the session-level intervals understate the variation between experiments. Doubling every U-Net width keeps the update speedup at 2.04 [2.01, 2.06] over three stall-free session pairs and lowers peak allocation by 56\%; at four times the width, both paths exceed the default allocator limit.

At batch 8 and $512\times512$, F completes all five sessions with a median update time of 197.30~ms and peak allocation of 12.758~GiB, whereas R runs out of memory during its first compiled update in every session, also when the allocator fraction is raised from 0.75 to 0.95 (Appendix~\ref{app:environment}). These outcomes depend on the allocator configuration and do not establish either implementation's maximum workload; the memory result is the 38--55\% reduction across jointly feasible workloads.

A separate device trace shows that F approximately halves cumulative kernel time. Reductions, already the largest kernel category in R (50\%), then account for about 65\% of F's kernel time. They include the batch statistics of midpoint normalization, which is identical in all paths, so faster normalization would complement factorization. Scaling speedup intervals are reported in Appendix~\ref{app:secondary}; Supplementary Table~S8 gives the trace summary.

\FloatBarrier
\subsection{Speedup attribution and Euclidean reference}\label{sec:attribution}
\begin{table}[!htbp]
\centering\small
\caption{Paired complete-model execution on the RTX A6000. Rg builds R's patch by gathering shifted slices; Mg keeps that patch but omits its ball mapping; E is a Euclidean network of the same topology and serves only as a cost reference. Ratios are R time divided by path time, with paired 95\% intervals over five sessions. R, Rg, Mg, and F share paired sessions; E rows come from separate paired R/F/E sessions whose R and F times differ from those shown by at most 1.5\%. Peak allocation includes initialization, compilation, and warmup; forward peaks are dominated by compilation workspace, which paths with the same convolutions share. These controls were added after the primary results.}
\label{tab:controls}
\setlength{\tabcolsep}{5pt}
\begin{tabular}{@{}llrcr@{}}
\toprule
\textbf{Workload} & \textbf{Path} & \textbf{Time (ms)} & \textbf{$t_{\mathrm R}/t$ [95\% interval]} & \textbf{Peak (GiB)}\\
\midrule
U-Net, batch 8, forward & R & 37.56 & -- & 1.400\\
 & Rg & 25.02 & $\speed{1.501}{1.500}{1.503}$ & 0.432\\
 & Mg & 18.80 & $\speed{1.995}{1.990}{2.001}$ & 0.432\\
 & F & 13.97 & $\speed{2.693}{2.685}{2.701}$ & 0.182\\
 & E & 3.09 & $\speed{12.18}{12.00}{12.37}$ & 0.182\\
\addlinespace
U-Net, batch 8, update & R & 105.18 & -- & 5.581\\
 & Rg & 92.41 & $\speed{1.138}{1.134}{1.142}$ & 4.458\\
 & Mg & 63.81 & $\speed{1.645}{1.636}{1.654}$ & 3.582\\
 & F & 50.63 & $\speed{2.070}{2.052}{2.089}$ & 2.364\\
 & E & 16.11 & $\speed{6.596}{6.481}{6.712}$ & 0.435\\
\addlinespace
ResNet-20, batch 128, update & R & 39.90 & -- & 1.762\\
 & Rg & 38.08 & $\speed{1.047}{1.035}{1.060}$ & 1.517\\
 & Mg & 28.37 & $\speed{1.406}{1.389}{1.423}$ & 1.193\\
 & F & 23.20 & $\speed{1.715}{1.641}{1.792}$ & 0.835\\
 & E & 6.88 & $\speed{5.542}{5.095}{6.027}$ & 0.237\\
\bottomrule
\end{tabular}
\end{table}
Table~\ref{tab:controls} tests whether the speedup depends on how R builds its patches. Under JAX 0.4.25, gathering shifted slices (Rg) is faster than R's one-hot patch convolution, by a factor of 1.14 for the U-Net update and 1.05 for the ResNet-20 update. Against this faster reference, F remains 1.82 [1.81, 1.83] times faster for the U-Net update and 1.64 [1.57, 1.71] times faster for the ResNet-20 update, with 47\% less peak memory for the U-Net update. Mg keeps the explicit patch but omits all pointwise ball geometry, so it approximates the cost of a lean implementation that materializes the patch; F is still 1.26 and 1.22 times faster than Mg for the two updates. Removing the ball mapping (Rg to Mg) thus accounts for a factor of 1.45 of the U-Net gain, and native convolution of the patch statistics \citep{chetlur2014} for a further 1.26.

\begin{table}[htbp]
\centering\small
\caption{Paired speedups of F over the explicit references on the RTX A6000 under two JAX versions: reference time divided by F time, with 95\% intervals over five sessions. The last column is the time of F relative to the Euclidean network E. JAX 0.4.25 values belong to Table~\ref{tab:controls}; all comparisons were added after the primary results.}
\label{tab:stacks}
\setlength{\tabcolsep}{4pt}
\begin{tabular}{@{}llcccc@{}}
\toprule
\textbf{Workload} & \textbf{JAX} & \textbf{Over R} & \textbf{Over Rg} & \textbf{Over Mg} & \textbf{F / E}\\
\midrule
U-Net, batch 8, forward & 0.4.25 & $\speed{2.69}{2.68}{2.70}$ & $\speed{1.79}{1.79}{1.80}$ & $\speed{1.35}{1.35}{1.35}$ & 4.5\\
 & 0.11.2 & $\speed{1.71}{1.70}{1.71}$ & $\speed{1.21}{1.21}{1.21}$ & $\speed{0.89}{0.89}{0.89}$ & 5.7\\
\addlinespace
U-Net, batch 8, update & 0.4.25 & $\speed{2.07}{2.05}{2.09}$ & $\speed{1.82}{1.81}{1.83}$ & $\speed{1.26}{1.24}{1.28}$ & 3.2\\
 & 0.11.2 & $\speed{1.67}{1.66}{1.67}$ & $\speed{1.52}{1.51}{1.52}$ & $\speed{1.26}{1.23}{1.28}$ & 5.7\\
\addlinespace
ResNet-20, batch 128, update & 0.4.25 & $\speed{1.71}{1.64}{1.79}$ & $\speed{1.64}{1.57}{1.71}$ & $\speed{1.22}{1.16}{1.28}$ & 3.2\\
 & 0.11.2 & $\speed{1.70}{1.70}{1.71}$ & $\speed{2.12}{2.11}{2.12}$ & $\speed{1.42}{1.41}{1.43}$ & 4.8\\
\bottomrule
\end{tabular}
\end{table}

Table~\ref{tab:stacks} repeats these comparisons with JAX 0.11.2 on an RTX A6000. For complete updates, F remains faster than every explicit reference: 1.52 and 2.12 times faster than Rg and 1.26 and 1.42 times faster than Mg for the U-Net and ResNet-20, with 45\% less peak memory than Rg for both updates. For the U-Net forward call, F is 1.21 times faster than Rg but slower than Mg. Device traces of the U-Net update show that reduction kernels, which include the batch statistics of the midpoint normalization shared by all paths, take about 16~ms longer per update under JAX 0.11.2 for both Rg and F, while F's convolution time is unchanged (Supplementary Tables~S17 and~S18). This common increase reduces F's relative advantage for the U-Net, whereas the gather-based reference slows down for ResNet-20. The Euclidean network shows the remaining cost of hyperbolic convolution: per update, F takes 3.2 times as long as E under JAX 0.4.25 and 4.8--5.7 times as long under JAX 0.11.2.

\begin{table}[htbp]
\centering\small
\caption{ResNet-20 complete update at batch 128 on V100 as the group widths $(8,16,32)$ are multiplied. Ratios are reference time divided by F's time with 95\% intervals over five paired sessions; peaks are process-lifetime peak live allocations. These post hoc measurements use JAX 0.4.25 and an allocator fraction of 0.95, and no session ran out of memory.}
\label{tab:width}
\setlength{\tabcolsep}{4pt}
\begin{tabular}{@{}rrccr@{}}
\toprule
\textbf{Width} & \textbf{Parameters} & \textbf{Over Rg} & \textbf{Over R} & \textbf{Peak R\,/\,Rg\,/\,F (GiB)}\\
\midrule
$\times1$ & 71,727 & $\speed{1.47}{1.19}{1.80}$ & $\speed{1.39}{1.17}{1.65}$ & 2.00\,/\,1.70\,/\,1.01\\
$\times2$ & 278,247 & $\speed{1.45}{1.35}{1.55}$ & $\speed{1.56}{1.41}{1.74}$ & 3.66\,/\,2.95\,/\,1.85\\
$\times4$ & 1,096,023 & $\speed{1.47}{1.38}{1.56}$ & $\speed{1.83}{1.73}{1.94}$ & 7.04\,/\,4.81\,/\,3.42\\
$\times8$ & 4,350,519 & $\speed{1.55}{1.49}{1.62}$ & $\speed{2.05}{1.91}{2.20}$ & 13.88\,/\,9.39\,/\,6.67\\
\bottomrule
\end{tabular}
\end{table}

Table~\ref{tab:width} tests whether the gain persists when channel contractions dominate the pointwise geometry. As the ResNet-20 widths grow eightfold, to 4.35 million parameters, F's advantage over Rg stays between 1.45 and 1.55 and its advantage over R grows from 1.39 to 2.05; F's peak allocation stays 29--41\% below Rg's and about 50\% below R's. The Rg ratios in this section are update-level measurements, and we did not train with Rg. Rg builds patches that are bitwise identical to R's, but its compiled gradients are not bitwise identical to R's (Supplementary Table~S19), so the training cohorts, which compare F with R, are not transferred to Rg.

\subsection{Official HypLL implementation}\label{sec:hypll}
To test whether the gain depends on our JAX reference, we implemented F in PyTorch, as a drop-in replacement included in the code release, for the \texttt{HConvolution2d} layer of HypLL 0.1.1, which Hyperbolic U-Net uses \citep{hypllcode,mishra2026}. With identical parameters, the factorized layer reproduces the official layer's outputs and gradients on S1--S3 and the stride-2 check to a relative L2 error of at most $10^{-14}$ in FP64 and $4.3\times10^{-6}$ in FP32. Separate stress tests expose shared curvature-gradient error near the ball boundary and output-map overflow; these agreement bounds do not cover those cases. On an RTX A6000, it accelerates forward-plus-gradient calls by factors of 1.43, 6.26, and 11.7 on S1--S3 and reduces their peak allocation by 3.8 to 9.0 times (Supplementary Table~S12). Replacing all 19 convolutions of the official Hyperbolic U-Net (depth 4, widths 8--128, 486,205 parameters), which uses iterative batch normalization rather than our midpoint normalization, makes a complete training step at batch 8 and $256\times256$ 1.41 [1.41, 1.42] times faster and halves its peak memory from 13.6 to 6.8~GiB. These peaks are steady-state allocator peaks and exclude compilation workspace; Table~\ref{tab:environment} lists the PyTorch environment.

\subsection{Training efficiency and task quality}\label{sec:training}
Factorization accelerates training updates in all four task and platform cohorts (Table~\ref{tab:trainingspeed}). Both implementations complete the same training budgets in the A6000 replications. The update speedups correspond to 50.4\% less compute time for ISIC16 and 43.3\% less for CIFAR-100. The improvement remains substantial when the clock includes compilation, evaluation, checkpointing, and other worker activity: elapsed time improves by approximately $1.65\times$ on both tasks. In the smaller V100 cohorts, both update intervals favor F, while the CIFAR-100 elapsed-time interval includes one. All 90 original training runs and the five repeated V100 CIFAR-100 R runs (95 in total) complete their fixed budgets without a recorded nonfinite loss, output, parameter, or Adam state. 

\begin{table}[htbp]
\centering\small
\caption{Training time per run and speedups for F relative to R. Times are mean minutes per run (R\,/\,F); speedups are geometric means of paired seed ratios with paired 95\% intervals, so they can differ from the quotient of the mean times. Bold speedup estimates have intervals entirely above one. Each row is analyzed separately. Elapsed worker time includes compilation, evaluation, checkpointing, and output writing. Three V100 CIFAR-100 F runs were resumed after an interruption, so their elapsed time includes a second start-up. V100 CIFAR-100 times are from the original R runs; the repeated R runs of Section~\ref{sec:training} give an update speedup of 1.342 [1.099, 1.639].}
\label{tab:trainingspeed}
\setlength{\tabcolsep}{4pt}
\begin{tabular}{@{}llrcccc@{}}
\toprule
 & & & \multicolumn{2}{c}{\textbf{Training updates}} & \multicolumn{2}{c}{\textbf{Elapsed worker time}}\\
\cmidrule(lr){4-5}\cmidrule(l){6-7}
\textbf{Task} & \textbf{GPU} & \textbf{Seeds} & \textbf{R\,/\,F (min)} & \textbf{Speedup} & \textbf{R\,/\,F (min)} & \textbf{Speedup}\\
\midrule
ISIC16 & V100 & 5 & 10.76\,/\,5.92 & $\bestspeed{1.821}{1.716}{1.934}$ & 25.06\,/\,19.02 & $\bestspeed{1.323}{1.208}{1.448}$\\
 & A6000 & 15 & 8.98\,/\,4.45 & $\bestspeed{2.017}{1.992}{2.042}$ & 12.44\,/\,7.58 & $\bestspeed{1.646}{1.610}{1.682}$\\
\addlinespace
CIFAR-100 & V100 & 5 & 30.65\,/\,22.91 & $\bestspeed{1.343}{1.192}{1.512}$ & 53.37\,/\,48.75 & $\speed{1.100}{0.984}{1.229}$\\
 & A6000 & 15 & 23.93\,/\,13.56 & $\bestspeed{1.764}{1.741}{1.787}$ & 35.37\,/\,21.38 & $\bestspeed{1.655}{1.624}{1.686}$\\
\bottomrule
\end{tabular}
\end{table}

\begin{figure}[htbp]
\centering
\includegraphics[width=\linewidth]{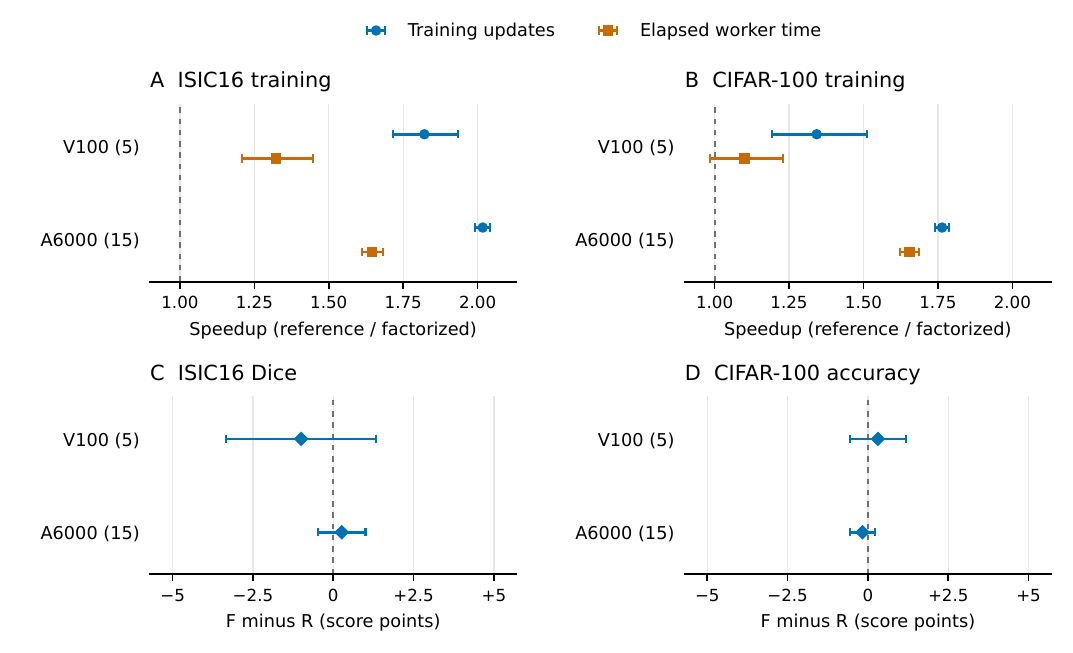}
\caption{Training efficiency and final task quality for F relative to R. Top panels show paired speedups for synchronized updates and elapsed worker time; bottom panels show paired quality differences on a 0--100 scale. Bars are unadjusted 95\% intervals. Dashed lines mark no change. V100 cohorts have five paired seeds and A6000 cohorts have 15. Cohorts remain separate; differences between platforms also reflect different seed sets and protocol versions.}
\label{fig:training}
\end{figure}

Task quality has no consistent direction of change (Table~\ref{tab:trainingoutcomes}). The A6000 intervals are narrower than the corresponding V100 intervals, and all four include zero. The A6000 estimates are $+0.26$ Dice points for ISIC16 and $-0.17$ accuracy points for CIFAR-100. The five original V100 CIFAR-100 R runs were affected by the stem-gradient deviation of Section~\ref{sec:agreement}; their stem weights grew far beyond those of every other run (Supplementary Table~S16). We repeated them post hoc, outside the registered protocol, with the weight normalization computed before the contraction, and use the repeated runs for this quality comparison; the original runs give $+2.21$ [$-0.81$, $5.24$]. The update speedup of F over the repeated runs, 1.34, equals that over the original runs, whose timings Table~\ref{tab:trainingspeed} keeps. These intervals describe seed variation and do not establish formal equivalence. Figure~\ref{fig:training} presents the quality estimates alongside the computational gains. Secondary checkpoint, duplicate-exclusion, and post hoc equivalence analyses are reported in Appendix~\ref{app:secondary}.

\begin{table}[htbp]
\centering\small
\caption{Final test quality on a 0--100 scale. Entries are mean $\pm$ standard deviation across seeds; differences and paired 95\% intervals are in score points. Individual seed outcomes are reported in Supplementary Tables~S4 and~S5. The V100 CIFAR-100 R entry uses the repeated runs described in Section~\ref{sec:training}.}
\label{tab:trainingoutcomes}
\setlength{\tabcolsep}{5pt}
\begin{tabular}{@{}llccc@{}}
\toprule
\textbf{Task / metric} & \textbf{GPU} & \textbf{R} & \textbf{F} & \textbf{F minus R [95\% interval]}\\
\midrule
ISIC16 Dice & V100 & $86.61\pm1.54$ & $85.61\pm1.40$ & $-1.00\;[-3.33,\;1.33]$\\
 & A6000 & $86.09\pm1.38$ & $86.35\pm1.02$ & $+0.26\;[-0.48,\;1.00]$\\
\addlinespace
CIFAR-100 top-1 & V100 & $49.83\pm0.64$ & $50.14\pm0.47$ & $+0.31\;[-0.56,\;1.18]$\\
 & A6000 & $49.71\pm0.60$ & $49.54\pm0.68$ & $-0.17\;[-0.55,\;0.21]$\\
\bottomrule
\end{tabular}
\end{table}

\section{Discussion}\label{sec:discussion}
The results show that a substantial part of the execution cost of Poincar\'e convolution comes from how its geometric score is assembled. The score needs directional products and patch energy, both of which can be computed without constructing a ball-valued receptive field at every spatial position. Expressing these quantities through native convolution and scalar reduction preserves the specified score in real arithmetic. The leading convolutional complexity remains unchanged, but the smaller intermediate representation reduces practical execution cost.

The strongest evidence is the combination of faster complete training and lower allocation during U-Net scaling. The A6000 replications show that the improvement survives beyond isolated kernels and remains visible in elapsed worker time. The V100 sweep shows a complementary benefit: peak allocation is 38--55\% lower across all jointly feasible workloads. These gains can matter when training time or memory limits the use of hyperbolic models. Their size depends on the workload. Normalization, reductions, and other unchanged operations limit overall speedup, and compilation-time autotuning can make allocation higher for a small forward workload. For the batch-8 U-Net and batch-128 ResNet-20 updates, a Euclidean network of the same topology still updates 2.7 to 5.7 times faster, depending on device and JAX version, so factorization reduces, but does not remove, the cost of hyperbolic convolution. Complete-model measurements also include the analogous dense score transformation, although the isolated dense benchmark does not establish a clear speedup.

Real-arithmetic equality and agreement with a floating-point reference do not guarantee identical optimization or accurate derivatives in every regime \citep{mishne2023}. In a constructed single-layer case at $c=10^{-3}$, an analytically zero curvature derivative evaluates in FP32 to $1.75$ for R, $1.625$ for F, and $0.026$ for T, and to about $10^{-9}$ in FP64; this diagnostic does not establish an effect on training. The learned ISIC16 curvature, which uses this derivative, shows no systematic difference between paths. Paired R and F trajectories separate within the first four epochs, but final values differ in both directions, with a mean F-minus-R difference of $+0.0035$ [$-0.0020$, $0.0089$] across 15 A6000 seed pairs (Supplementary Table~S10 and Figure~S1); these differences are not associated with the paired Dice differences (Spearman $\rho=0.00$, $p=1.0$). At trained states, FP32 gradients of F agree with an FP64 reference at least as closely as those of the reference implementation. The training cohorts show no consistent quality difference. Final quality differences change sign between cohorts, with small gains and losses remaining plausible in the A6000 estimates.

Several limits define the scope of the measurements. The training cohorts use the pinned JAX 0.4.25 stack on two GPUs and measure F against R; the gather-based and Euclidean controls, the JAX 0.11.2 repetition, and the PyTorch cross-check are update-level benchmarks rather than training runs. On the RTX A6000, against the faster of the two explicit implementations that map patches to the ball, the update speedup is 1.6--1.8$\times$ under JAX 0.4.25 and 1.5--1.7$\times$ under JAX 0.11.2; against the explicit implementation that skips this mapping, it is 1.2--1.3$\times$ and 1.3--1.4$\times$, respectively. The U-Net forward call shows that the benefit of native convolution of the statistics depends on the compiler. The scaling study uses repeated synthetic batches and newly initialized states, so it measures device execution and allocation rather than a full data pipeline. The cohorts do not isolate a causal hardware effect. Current-batch normalization also makes test results depend on batch composition, and the ISIC16 split supports a retrospective benchmark comparison rather than claims about patient-level generalization. The widest model we timed has 4.35 million parameters. Further work can establish how the gain behaves with other convolution backends, larger architectures, and the Lorentz layers discussed next.

The factorization extends beyond HNN++ when two conditions hold: each patch is a fixed-scale concatenation of per-location vectors in one coordinate system, and the layer depends on the patch only through its squared norm and inner products with learned vectors. M\"obius-addition concatenation \citep{ganea2018} violates the first condition, whereas Lorentz direct concatenation satisfies it. On the hyperboloid $\{x:-x_0^2+\norm{x_s}^2=-1/c\}$, concatenating the points of a patch with space scale $s$ gives
\begin{equation}
 x_{p,0}=\Bigl(\frac1c+s^2\sum_{q\in\mathcal P_p}\norm{x_{q,s}}^2\Bigr)^{1/2},\qquad
 x_{p,s}=s\,[\,x_{q,s}\,]_{q\in\mathcal P_p}.
 \label{eq:lorentzcat}
\end{equation}

By \eqref{eq:lorentzcat}, the time coordinate is obtained elementwise from a single-channel box filter of the squared space norms. A Lorentz fully connected layer whose space output is an affine map of $(x_{p,0},x_{p,s})$, as in HCNN \citep{bdeir2024}, then equals a native convolution of the space channels plus a weighted copy of $x_{p,0}$. The public implementations of these Lorentz convolutions \citep{bdeir2024,klis2026}, like those of HypLL and Poincar\'e ResNet \citep{hypllcode,poincareresnetcode}, build explicit patches by unfolding. We have not evaluated this extension.

\section{Conclusion}
Poincar\'e convolution can be evaluated from patch energy and directional products without materializing ball-valued patches. The factorization preserves the specified clipped HNN++ score in real arithmetic, agrees closely with the reference in floating point, and makes complete training updates faster and less memory-intensive than both the published execution order and a gather-based explicit implementation in the tested models, devices, and software stacks. Normalization and other unchanged operations now account for most of the remaining cost, so faster hyperbolic normalization would complement factorization.

\section*{CRediT authorship contribution statement}

\textbf{Luk\'a\v{s} Jochymek:}
Conceptualization, Methodology, Investigation, Formal analysis, Software, Writing -- original draft, Writing -- review \& editing.
\textbf{Mark\'eta Va\v{s}inkov\'a:} Data curation, Writing -- original draft, Visualization, Writing -- review \& editing.
\textbf{Petr Gajdo\v{s}:} Validation, Supervision, Project administration, Funding acquisition,
Writing -- review \& editing.

All authors reviewed and approved the final manuscript.

\section*{Declaration of competing interest}
The authors declare that they have no known competing financial interests or personal relationships that could have appeared to influence the work reported in this paper.

\section*{Acknowledgements}
This work was supported by the Center for Artificial Intelligence and Quantum Computing in System Brain Research (CLARA) (101136607-02), the project “Research Platform for Digital Transformation and Society 5.0” (CZ.02.01.01/00/23\_021/0012599), the Internal Grant Agency of VSB-TUO, Processing and Advanced Analysis of Biomedical Data XI of the Czech Republic (SP2026/008).

\section*{Code and data availability}
The source code, experimental configurations, tests, raw timing and training logs, and the scripts used to generate all tables and figures are publicly available in the ExPoConv repository (\url{https://github.com/lowoncuties/ExPoConv}). The exact version used
in this study is archived on Zenodo \citep{jochymek2026software}.
No new data were collected. The experiments use the publicly available
ISIC~2016 skin lesion segmentation dataset \citep{gutman2016} and the
CIFAR-100 dataset \citep{krizhevsky2009}, which can be obtained from their original providers under their respective terms of use.

\appendix

\numberwithin{equation}{section}
\numberwithin{table}{section}
\numberwithin{figure}{section}

\section{Implementation details}\label{app:implementation}
\subsection{Direct tangent output and identity}\label{app:direct}\label{app:proof}

At one output location, let $v\in\R^{C_{\mathrm{out}}}$ be the score vector, define $b=\sinh(\sqrt c\,v)$ componentwise, and set the scalar $S=\norm b$. The HNN++ reconstruction and its origin-tangent image before output clipping are

\begin{equation}
 y=\frac{b}{\sqrt c(1+\sqrt{1+S^2})},\qquad
 t_{\mathrm{out}}=\frac{\asinh(S)}{2\sqrt c}\frac{b}{S}.
 \label{eq:directoutput}
\end{equation}

The second expression has the smooth extension $t_{\mathrm{out}}=v/2+O(\norm{v}^3)$ and Jacobian $I/2$ at $v=0$. R and F reconstruct $y$, project it to the prescribed ball radius, and apply the logarithmic map. T computes the tangent expression directly and applies the corresponding cap. With scale one at a zero vector, these projections are
\begin{equation}
 \Pi_{\B}^{c,\epsilon}(x)=x\min\!\left(1,\frac{m}{\sqrt c\norm{x}}\right),\qquad
 \Pi_{T}^{c,\epsilon}(t)=t\min\!\left(1,\frac{\atanh(m)}{\sqrt c\norm{t}}\right).
 \label{eq:outputclip}
\end{equation}
Thus R and F return $\log_0^c(\Pi_{\B}^{c,\epsilon}(y))$, while T returns $\Pi_T^{c,\epsilon}(t_{\mathrm{out}})$. The two expressions share the same output boundary in real arithmetic.

For $S>0$, Eq.~\eqref{eq:directoutput} follows from

\begin{equation}
 \atanh\!\left(\frac{S}{1+\sqrt{1+S^2}}\right)=\frac12\asinh(S)
\end{equation}

and the shared direction of $y$ and $b$. Series expansion gives the smooth zero extension and $D t_{\mathrm{out}}(0)=I/2$. For one output channel, $\asinh(\sinh(\sqrt c\,v))=\sqrt c\,v$ in real arithmetic, so $t_{\mathrm{out}}=v/2$ before projection. These real-arithmetic identities do not imply bitwise floating-point identity or identical optimization trajectories.

\FloatBarrier
\subsection{Scaled output evaluation and model composition}
For large scores the implementation avoids materializing $\sinh(\sqrt c\,v)$. Let
\begin{equation}
 d=\sqrt c\,v,\qquad
 M=\operatorname{stopgrad}\!\left(\max_j |d_j|\right),\qquad
 s_j=\frac12\left(e^{d_j-M}-e^{-d_j-M}\right),
 \label{eq:scaledreadout}
\end{equation}
and set $Q=\norm{s}$ and $\ell=M+\log Q$. When $M>20$, the two outputs in Eq.~\eqref{eq:directoutput} are evaluated as
\begin{equation}
 h(\ell)=
 \begin{cases}
  \ell+\log\!\left(1+\sqrt{1+e^{-2\ell}}\right),&\ell\geq0,\\
  \asinh(e^\ell),&\ell<0,
 \end{cases}
 \quad
 r=\frac{h(\ell)}{2\sqrt c},\quad
 t_{\mathrm{out}}=r\frac{s}{Q},\quad
 y=\frac{\tanh(\sqrt c\,r)}{\sqrt c}\frac{s}{Q}.
 \label{eq:scaledreadoutoutputs}
\end{equation}
The direct expression is used when $M\leq20$; thus equality at the threshold selects the direct branch. The maximum in Eq.~\eqref{eq:scaledreadout} is stopped only for differentiation of the scaling device, not for differentiation of the scores themselves.

Following the established exp--log lifting of \citet{ganea2018}, for a supported ball operator $G$ let $\widetilde G=\log_0^c\circ G\circ\exp_0^c$. Matching exponential and logarithmic maps cancel in exact compositions with the same curvature, provided no intervening projection is removed. Tangent ReLU is then ordinary ReLU. For tangent vectors $t_1\in\R^m$ and $t_2\in\R^n$, beta concatenation becomes

\begin{equation}
 [\,(\beta_{m+n}/\beta_m)t_1\ ;\ (\beta_{m+n}/\beta_n)t_2\,].
 \label{eq:betaskip}
\end{equation}

The controlled models retain midpoint normalization, ordered M\"obius residuals, pooling, and transpose decoding in their hyperbolic forms. Transfers between different curvatures and the order of M\"obius additions remain explicit. Transpose decoding uses a fully connected operation followed by folding.

\FloatBarrier
\subsection{Derivatives and small-argument evaluation}

Projection uses radius $(1-\epsilon)/\sqrt c$, with $\epsilon=0.004$ in FP32 and $10^{-5}$ in FP64. At the origin the projection Jacobian is the identity. For $f(x)=\norm{x}$ and $x\ne0$, the norm gradient is $x/\norm{x}$, whereas its Jacobian--vector product in direction $\dot x$ is

\begin{equation}
 Df(x)[\dot x]=\frac{\langle x,\dot x\rangle}{\norm{x}}.
 \label{eq:normjvp}
\end{equation}

The implementation evaluates Eq.~\eqref{eq:normjvp} without reciprocal-square intermediates and selects zero at $x=0$, where the norm is nondifferentiable; this is a derivative convention. At an exact radius cap the maps are also nondifferentiable. The explicit ball and tangent projection helpers use a strict outside test, so machine equality selects their identity branch. By contrast, the patch-statistics score cap obtains its normalized squared radius with \texttt{jnp.minimum}; under the pinned JAX 0.4.25 autodifferentiation convention, equality assigns a derivative of $1/2$ with respect to its first scalar argument. These are implementation tie conventions rather than unique analytic derivatives, and constructed-cap rows are diagnostics outside the smooth acceptance gate. Near zero, the implementation uses the sixth-order Taylor approximations
\begin{align}
 \frac{\sinh x}{x}&\approx1+\frac{x^2}{6}+\frac{x^4}{120}+\frac{x^6}{5040}, &
 \frac{\asinh x}{x}&\approx1-\frac{x^2}{6}+\frac{3x^4}{40}-\frac{5x^6}{112},\nonumber\\
 \frac{\atanh x}{x}&\approx1+\frac{x^2}{3}+\frac{x^4}{5}+\frac{x^6}{7}, &
 \frac{\tanh x}{x}&\approx1-\frac{x^2}{3}+\frac{2x^4}{15}-\frac{17x^6}{315},\label{eq:series}\\
 \tanh^2 x&\approx x^2-\frac{2x^4}{3}+\frac{17x^6}{45}.&&\nonumber
\end{align}
The series branch is selected for $|x|\leq10^{-3}$ in FP32 and $|x|\leq10^{-6}$ in FP64; functions evaluated from $x^2$ use the corresponding squared threshold. A high-precision oracle evaluating the FP32 contract retains the FP32 clipping margin so that it evaluates the same clipped function.

For the filter factor $\bar g=\max(g,10^{-15})$, $g=\norm z$, the pinned JAX autodifferentiation convention gives $\partial\bar g/\partial g=0$, $1/2$, and $1$ below, at, and above the floor, respectively. Combined with the zero derivative selected for $g$ at $z=0$, this specifies the implementation derivative at both exceptional points. The floor itself remains a mathematical kink; no unique analytic derivative is claimed there.

Figure~\ref{fig:overview} summarizes the two execution changes. All paths retain the same normalization, curvature parameterization, and initialization. A learned curvature then evolves separately during each training run. Source and output projections are geometric operations; the U-Net optimizer's global gradient-norm clip is a separate operation.

\section{Experimental recipes}\label{app:protocol}

The 15 A6000 replication seeds are listed in Supplementary Table~S5. R and F use the same seed, initialization, and data order within each pair.

\FloatBarrier
\subsection{ISIC16 Poincar\'e U-Net}

The controlled U-Net has depth four; channel widths $(8,16,32,64,128)$; two convolutions per block; current-batch Poincar\'e midpoint normalization ($\epsilon=10^{-6}$, no running state); tangent max pooling; beta-concatenated skips; FC-plus-fold transpose decoding; and a two-channel output. It has 486,941 trainable scalars and 19 affected spatial Poincar\'e convolutions; F also changes dense score evaluation in the fully connected component of each transpose-decoder operation. Its learned curvature is $c=\operatorname{softplus}(\theta)$ with $\theta_0=0.1+\log[-\operatorname{expm1}(-0.1)]$; there is no additive curvature floor or clamp.

ISIC16 contains 900 official training and 379 official test image--mask pairs. Seed 1729 partitions the official training set into 810 training and 90 development examples. Each image is padded on the bottom and right to a square using its minimum intensity and resized bicubically to $256\times256$ with Pillow 11.3; masks are zero-padded and resized with nearest-neighbor interpolation. Inputs are scaled to $[0,1]$ without augmentation. Predictions are hard argmax labels, with class 1 as foreground. Per-image Dice is $2TP/(2TP+FP+FN)$ and IoU is $TP/(TP+FP+FN)$; a zero denominator receives one, and endpoints average over images.

Each epoch applies a deterministic seed- and epoch-specific permutation to all 810 examples, producing 101 batches of eight and one final batch of two; no example is dropped, padded, or duplicated. Separate compiled update functions handle the two batch sizes. Adam \citep{kingma2015} uses $\beta_1=0.9$, $\beta_2=0.999$, $\epsilon=10^{-8}$, learning rate $10^{-3}$, and no schedule. The global norm of the loss gradient is clipped at 1 before coupled $10^{-4}$ L2 is added and the Adam moments are updated; the L2 term is not part of the reported scalar loss.

For the two-class soft Dice term \citep{milletari2016}, probabilities and one-hot targets are first summed jointly over batch and spatial dimensions. For class $k$, with intersection $I_k=\sum p_k y_k$ and denominator $D_k=\sum(p_k+y_k)$, the score is $(2I_k+10^{-6})/\max(D_k+10^{-6},10^{-7})$. An absent class contributes zero loss, after which the two class losses are averaged. The focal term \citep{lin2017} uses stable one-vs-all binary cross-entropy, $\alpha=0.25$ and $\gamma=2$; it is averaged over batch and spatial dimensions separately for each class and then summed over the two classes. The total scalar objective is the unit-weight sum of the Dice and focal terms. Evaluation batch size is eight: the development, official-test, and 376-image filtered sets end with batches of two, three, and eight, respectively.

\FloatBarrier
\subsection{CIFAR-100 Poincar\'e ResNet-20}

The compact ResNet-20 has three residual groups, three blocks per group, widths $(8,16,32)$, ordered function-then-residual M\"obius addition, and one Poincar\'e classifier. It has 71,727 trainable scalars and 21 affected spatial convolutions; F also changes the classifier's dense score evaluation. Midpoint normalization uses current-batch statistics ($\epsilon=10^{-6}$, no running state), and curvature is fixed at $c=0.1$.

Seed 1729 partitions the 50,000 official training images into 45,000 training and 5,000 development examples without train--development byte-identical overlap; the 10,000 official test images remain primary. Inputs are scaled to $[-1,1]$. Deterministic example-indexed augmentation uses random crops with four-pixel padding and horizontal flips of probability 0.5. Adam uses the U-Net coefficients, learning rate, and coupled L2, but no learning-rate schedule or gradient clipping. The scalar objective is the batch mean of the selected negative log-softmax probability, with label smoothing exactly zero. Batch size 128 for 100 epochs gives 35,200 updates, including one retained 72-example remainder per epoch. Development and test evaluation use manifest order and batch size 128. Of ten official-test inputs byte-identical to official-training inputs, four duplicate groups share the fine label and six conflict.

\FloatBarrier
\subsection{Controlled choices}\label{app:controlled}

Relative to the inspected U-Net branch, this study selects midpoint rather than iterative Fr\'echet normalization, implements the declared Dice--focal objective, uses one model-wide learned curvature per path, and uses fixed-rate Adam without validation-driven scheduling. The ResNet uses the compact published $(8,16,32)$ topology \citep{spengler2023,spengler2025adversarial} but evaluates the published HNN++ conformal-factor quotient rather than an inspected upstream multiplication. The upstream revisions and experimental configurations are recorded in the experiment metadata. These settings define matched R/F/T testbeds; none is attributed to F.

\FloatBarrier
\subsection{Software environment}\label{app:environment}
Table~\ref{tab:environment} gives the measured runtime stacks. Both platforms use the CUDA-11 build of \texttt{jaxlib} and pinned CUDA-11.8 libraries, irrespective of the driver's advertised CUDA compatibility. Timed tensors are FP32; JAX 64-bit support is enabled for validation. Matrix-multiplication precision is \texttt{highest}; reduced-precision reduction and TF32 are disabled; \texttt{XLA\_PYTHON\_CLIENT\_PREALLOCATE=false}; and \texttt{XLA\_FLAGS} and \texttt{NVIDIA\_TF32\_OVERRIDE} are unset. The system compiler is not used for the prebuilt \texttt{jaxlib}; no independent XLA revision was recorded.

\begin{table}[htbp]
\centering\footnotesize
\caption{Runtime environment. Package versions are shared unless stated otherwise. Driver CUDA compatibility does not identify the runtime used by the experiment.}
\label{tab:environment}
\begin{tabularx}{\linewidth}{@{}p{0.29\linewidth}YY@{}}
\toprule
\textbf{Component} & \textbf{V100} & \textbf{RTX A6000}\\
\midrule
Python / JAX & 3.11.16 / 0.4.25 & Same\\
\texttt{jaxlib} & \texttt{0.4.25+cuda11.cudnn86} & Same\\
NumPy / SciPy & 1.26.4 / 1.11.4 & Same\\
CUDA runtime & 11.8.89 & Same\\
cuDNN / cuBLAS & 8.6.0.163 / 11.11.3.6 & Same\\
NVIDIA driver & 520.61.05 & 560.35.03\\
Driver CUDA compatibility & 11.8 & 12.6\\
JAX 0.11.2 controls (Section~\ref{sec:attribution}) & -- & Python 3.12.3, JAX and \texttt{jaxlib} 0.11.2, CUDA-12 plugin, cuDNN 9.27.0\\
PyTorch check (Section~\ref{sec:hypll}) & -- & Python 3.10.21, PyTorch 2.4.0, CUDA 12.1, cuDNN 9.1.0, HypLL 0.1.1\\
\bottomrule
\end{tabularx}
\end{table}

\subsection{Measurement details}\label{app:measurement}
The V100 measurements of E and of the twice-wider U-Net ran concurrently on the two identical V100s of one host: E on the second device and R and F on the device of the other V100 measurements. Sessions that overlapped a logged GPU driver stall are excluded under a rule fixed before their latencies were inspected: one of five E sessions for the batch-8 update of Table~\ref{tab:unettime}, three of five E sessions at batch 1 and at batch 2 in Figure~\ref{fig:scaling}, and two of five width pairs. Including them changes no conclusion.

At batch 8 and $512\times512$, XLA's compiled buffer allocation for R's update is 23.0~GiB, below the 23.81~GiB cap; the update fails because earlier allocations already occupy part of the allocator pool. With the allocator fraction raised to 0.95, F again completes every session at 12.758~GiB, whereas R fails with a 28.4~GiB compiled allocation, consistent with XLA selecting convolution algorithms with larger workspaces when more memory is available. At $512\times512$ and batch 16, both paths fail, and a $256\times256$ control under the raised fraction reproduces the peak allocations of Table~\ref{tab:unettime}.

The ResNet-20 width sweep of Table~\ref{tab:width} ran on the second V100 while the other V100 of the same host trained the repeated CIFAR-100 R runs of Section~\ref{sec:training}; no session overlapped a logged driver stall.

For one V100 ISIC16 seed, R and T ran at the code revision preceding that of all other runs; the two revisions differ only in the launcher's support for F.

A separate within-project rerun on the original V100 stack gave median update times of 123.236~ms for R and 73.883~ms for F across five paired sessions, with paired speedup $1.745$ [1.557, 1.956]. It checks the timing workflow without repeating training or the peak-allocation study and remains separate from the main estimates.

The JAX 0.11.2 measurements of Section~\ref{sec:attribution} use the CUDA-12 plugin with cuDNN 9.27 on one RTX A6000 of the training hosts, with the allocator fraction of 0.95. This A6000 is a second device of the same model as that of Table~\ref{tab:controls}; the JAX 0.4.25 traces were repeated on it. A co-resident training process was suspended during timing and tracing; it launched no work but held about 16.6~GiB, so the allocator limit was computed from the reduced free memory. The measurements supersede an earlier single-workload check with the same versions, which gave the same R/F update ratio (1.667 against 1.668). Both training cohorts retain JAX 0.4.25 as their recorded software environment.

\section{Additional comparisons}\label{app:secondary}
\subsection{Direct tangent output}
Table~\ref{tab:directcomparison} isolates the additional output simplification by comparing F directly with T on V100. T gives small forward gains on the larger convolutions but slows their forward-plus-gradient calls. The complete-update and training intervals show no clear additional timing benefit. At batch eight, complete-update peak allocation is 2.721~GiB for F and 3.409~GiB for T. This supports F, which keeps the reference output calculation, as the primary implementation. Supplementary Tables~S1--S3 retain the full R/F/T comparisons.

\begin{table}[htbp]
\centering\small
\caption{Paired F-to-T latency ratios on V100. Values above one favor T; values below one favor F. Brackets are unadjusted 95\% intervals. Bold estimates have intervals that exclude one in either direction. These are directly reported paired ratios, not quotients of reference-relative estimates.}
\label{tab:directcomparison}
\begin{tabular}{@{}llc@{}}
\toprule
\textbf{Workload} & \textbf{Boundary} & \textbf{$t_{\mathrm F}/t_{\mathrm T}$ [95\% interval]}\\
\midrule
S2 & Forward & $\bestspeed{1.055}{1.048}{1.062}$\\
S3 & Forward & $\bestspeed{1.055}{1.009}{1.103}$\\
S2 & Forward + gradient & $\bestspeed{0.833}{0.814}{0.853}$\\
S3 & Forward + gradient & $\bestspeed{0.882}{0.863}{0.900}$\\
\addlinespace
U-Net, batch 8 & Complete update & $\speed{0.990}{0.923}{1.062}$\\
ISIC16 U-Net & Training updates & $\speed{0.965}{0.840}{1.108}$\\
CIFAR-100 ResNet-20 & Training updates & $\speed{0.990}{0.838}{1.170}$\\
\bottomrule
\end{tabular}
\end{table}

\subsection{Sensitivity analyses}
Table~\ref{tab:sensitivitysummary} summarizes the F-minus-R contrasts after duplicate exclusion and at development-selected checkpoints. Duplicate exclusion changes the composition of normalization batches, so its effect includes that change. Checkpoint selection uses development data only. All secondary intervals include zero; V100 CIFAR-100 contrasts use the repeated R runs. Supplementary Tables~S4--S7 retain individual seeds, original precision, further metrics, and the T outcomes.

\begin{table}[htbp]
\centering\small
\caption{Secondary quality differences for F minus R on the native 0--1 metric scales. ISIC16 uses foreground Dice and CIFAR-100 uses top-1 accuracy. Brackets show unadjusted paired 95\% intervals across training seeds.}
\label{tab:sensitivitysummary}
\setlength{\tabcolsep}{4pt}
\begin{tabular}{@{}lllc@{}}
\toprule
\textbf{Task} & \textbf{GPU} & \textbf{Endpoint} & \textbf{F minus R [95\% interval]}\\
\midrule
ISIC16 & V100 & Duplicate-excluded & $-0.01050\;[-0.02872,\;0.00773]$\\
 & A6000 & Duplicate-excluded & $+0.00292\;[-0.00331,\;0.00915]$\\
 & V100 & Development-selected & $-0.00683\;[-0.01380,\;0.00015]$\\
 & A6000 & Development-selected & $-0.00229\;[-0.00684,\;0.00226]$\\
\addlinespace
CIFAR-100 & V100 & Duplicate-excluded & $+0.00146\;[-0.00653,\;0.00946]$\\
 & A6000 & Duplicate-excluded & $-0.00105\;[-0.00503,\;0.00292]$\\
 & V100 & Development-selected & $-0.00008\;[-0.01418,\;0.01402]$\\
 & A6000 & Development-selected & $-0.00117\;[-0.00541,\;0.00308]$\\
\bottomrule
\end{tabular}
\end{table}

Post hoc two one-sided tests \citep{schuirmann1987} of the primary contrasts in Table~\ref{tab:trainingoutcomes} use margins of 1 Dice point and 0.5 accuracy points that were not prespecified. At an unadjusted level of 0.05, equivalence within these margins is met in both A6000 cohorts (ISIC16 $p=0.026$; CIFAR-100 $p=0.043$) and in neither five-seed V100 cohort ($p=0.50$ and $0.29$). The smallest margins that would be met, given as the 90\% interval bounds rounded up to 0.01, are 0.88 Dice and 0.49 accuracy points on the A6000 and 2.80 Dice and 0.99 accuracy points on the V100.

\subsection{Scaling estimates}
Across the six jointly completed scaling configurations, the smallest paired speedup is $1.294$ [1.058, 1.582] at batch 1 and $256\times256$; the largest is $2.103$ [2.073, 2.132] at batch 16 and $256\times256$. The shared batch-8, $256\times256$ speedup is $1.935$ [1.907, 1.964]. The recorded allocator limit is 25,567,444,992 bytes (23.81~GiB). Supplementary Table~S8 gives the descriptive device-trace summary.

\begin{samepage}
\section*{Declaration of generative AI and AI-assisted technologies in the manuscript preparation process}
During the preparation of this manuscript, the authors used OpenAI Codex and Anthropic Claude to assist with language editing and improve the clarity, readability, and academic style of the text. The authors reviewed and edited all suggested revisions and take full responsibility for the content of the published article.

\end{samepage}

\printbibliography

\end{document}

\maketitle
These tables retain the reported detailed comparisons, individual seed outcomes, secondary endpoints, and device trace, and report elementwise agreement, learned-curvature, and official-HypLL comparison results referenced in the main article. R is the reference implementation, F is score factorization with the reference output calculation, and T additionally uses direct tangent output. The main article specifies the timing boundaries, statistical analysis, and experimental recipes.

\section{Detailed results}
\FloatBarrier
\subsection{Direct-tangent ablation}
Tables~\ref{tab:operatorsfull}--\ref{tab:trainingoutcomesfull} report the direct-tangent ablation alongside R and F. For any paths X and Y, $t_{\mathrm X}/t_{\mathrm Y}$ denotes their paired timing ratio; a value above one favors Y.

\begin{table}[htbp]

\centering\footnotesize

\caption{Detailed paired operator speedups. Bold marks the larger point estimate in each row, without implying a significant difference between F and T. Each ratio is reference latency divided by candidate latency; values above one favor the candidate. ``Fwd+grad'' includes forward evaluation and reverse-mode gradients but no optimizer update.}

\label{tab:operatorsfull}

\begin{tabular}{llcc}
\toprule
\textbf{Boundary} & \textbf{Mode} & \textbf{$t_{\mathrm R}/t_{\mathrm F}$} & \textbf{$t_{\mathrm R}/t_{\mathrm T}$}\\
\midrule
S1 ($16^2$, $256\!\to\!256$) & forward & 1.833 [1.740, 1.932] & \best{1.864} [1.825, 1.904]\\
   & fwd+grad & \best{2.070} [1.940, 2.210] & 1.970 [1.769, 2.194]\\
S2 ($256^2$, $32\!\to\!32$) & forward & 9.581 [9.478, 9.685] & \best{10.110} [10.011, 10.210]\\
   & fwd+grad & \best{3.382} [3.310, 3.457] & 2.818 [2.755, 2.883]\\
S3 ($256^2$, $96\!\to\!32$) & forward & 25.810 [25.027, 26.618] & \best{27.228} [26.683, 27.783]\\
   & fwd+grad & \best{5.668} [5.585, 5.752] & 4.996 [4.912, 5.082]\\
Conv--BN--ReLU & forward & \best{1.166} [1.024, 1.328] & 1.116 [1.002, 1.244]\\
   & fwd+grad & 1.052 [0.977, 1.131] & \best{1.077} [1.032, 1.124]\\
Dense stack (4) & forward & 1.090 [0.935, 1.270] & \best{1.142} [1.004, 1.299]\\
   & fwd+grad & \best{1.102} [0.998, 1.217] & 0.993 [0.920, 1.071]\\
\bottomrule
\end{tabular}

\end{table}

Direct F-to-T comparisons favor T slightly in forward execution on S2 and S3: the $t_{\mathrm F}/t_{\mathrm T}$ latency ratios are $1.055$ [1.048, 1.062] and $1.055$ [1.009, 1.103], where values above one favor T. For forward plus gradient they fall to $0.833$ [0.814, 0.853] and $0.882$ [0.863, 0.900]. Direct tangent readout therefore speeds these forward calls but slows their gradient boundaries.

\begin{table}[htbp]

\centering\footnotesize

\caption{Complete U-Net latency and peak live allocation on V100. Bold marks the lowest time and allocation, and the largest paired speedup within each workload; ties are retained. The paired reference/candidate timing ratio can differ from the quotient of displayed median latencies.}

\label{tab:unettimefull}

\begin{tabular}{llrrr}
\toprule
\textbf{Boundary} & \textbf{Variant} & \textbf{Latency (ms)} & \textbf{Speedup [95\% interval]} & \textbf{Peak live (GiB)}\\
\midrule
Batch-1 forward & R & 9.831 & -- & \best{0.237}\\
 & F & \best{5.143} & 1.864 [1.787, 1.944] & 0.395\\
 & T & 5.222 & \best{1.890} [1.854, 1.926] & 0.395\\
\midrule
Batch-8 forward & R & 48.438 & -- & 0.885\\
 & F & 17.858 & 2.687 [2.608, 2.768] & \best{0.619}\\
 & T & \best{17.655} & \best{2.741} [2.678, 2.804] & \best{0.619}\\
\midrule
Batch-8 update & R & 124.062 & -- & 5.955\\
 & F & \best{64.325} & \best{1.923} [1.755, 2.107] & \best{2.721}\\
 & T & 65.322 & 1.904 [1.830, 1.981] & 3.409\\
\bottomrule
\end{tabular}

\end{table}

At batch eight, process-lifetime peak live allocation is 54.3\% lower than the reference for factorized execution and 42.8\% lower for direct tangent output. The paired complete update ratio $t_{\mathrm F}/t_{\mathrm T}$ is 0.990 [0.923, 1.062]. Direct tangent output therefore gives no clear update-time improvement over factorization and uses more peak allocation for the complete update.

\begin{table}[htbp]

\centering\footnotesize

\setlength{\tabcolsep}{4pt}

\caption{Detailed final outcomes and synchronized training update speedups for the original five-seed V100 cohorts. Differences are candidate minus R. CIFAR-100 R quality uses the repeated runs of main-text Section~5.6; update speedups use the original runs. Bold marks the highest observed mean and the largest speedup per task; quality emphasis is descriptive.}

\label{tab:trainingoutcomesfull}

\begin{tabular}{llccc}
\toprule
\textbf{Task / metric} & \textbf{Path} & \textbf{Mean $\pm$ SD} & \textbf{Difference [95\% interval]} & \textbf{Update speedup}\\
\midrule
ISIC16 Dice & R & $\mathbf{0.86609}\pm0.01544$ & -- & --\\
 & F & $0.85607\pm0.01400$ & $-0.01002$ [$-0.03335$, $0.01332$] & \best{1.821} [1.716, 1.934]\\
 & T & $0.86593\pm0.00574$ & $-0.00016$ [$-0.01981$, $0.01950$] & 1.758 [1.593, 1.940]\\
\midrule
CIFAR-100 top-1 & R & $0.49832\pm0.00641$ & -- & --\\
 & F & $\mathbf{0.50144}\pm0.00465$ & $+0.00312$ [$-0.00559$, $0.01183$] & \best{1.343} [1.192, 1.512]\\
 & T & $0.49238\pm0.00658$ & $-0.00594$ [$-0.01269$, $0.00081$] & 1.329 [1.167, 1.513]\\
\bottomrule
\end{tabular}

\end{table}

The paired training update ratios $t_{\mathrm F}/t_{\mathrm T}$ are 0.965 [0.840, 1.108] for U-Net and 0.990 [0.838, 1.170] for ResNet-20. Neither identifies an additional training compute benefit from direct tangent output.

\FloatBarrier
\subsection{Individual final training outcomes}

Table~\ref{tab:seedoutcomes} lists the original V100 endpoints with the highest observed score in each seed shown in bold; Table~\ref{tab:replicationseeds} lists the A6000 replication pairs. The main article displays the primary F-minus-R differences in all four cohorts.

\begin{table}[htbp]

\centering\footnotesize

\setlength{\tabcolsep}{4pt}

\caption{Individual final official-test endpoints in the original V100 cohorts. Differences are candidate minus R. Bold marks the highest observed score within each seed. CIFAR-100 R values come from the repeated runs that remove the miscompiled stem gradient (main-text Section~5.6); the original runs reached 0.4878, 0.4939, 0.4340, 0.4963, and 0.4845.}

\label{tab:seedoutcomes}

\begin{tabular}{llrrrrr}
\toprule
\textbf{Task} & \textbf{Seed} & \textbf{R} & \textbf{F} & \textbf{T} & \textbf{F--R} & \textbf{T--R}\\
\midrule
ISIC16 Dice & 11  & \best{0.87927} & 0.86861 & 0.87532 & $-0.01066$ & $-0.00395$\\
 & 29  & \best{0.87179} & 0.83465 & 0.86054 & $-0.03714$ & $-0.01125$\\
 & 47  & 0.83940 & 0.85263 & \best{0.86709} & $+0.01323$ & $+0.02768$\\
 & 71  & \best{0.87158} & 0.85589 & 0.86329 & $-0.01569$ & $-0.00829$\\
 & 101 & 0.86841 & \best{0.86858} & 0.86343 & $+0.00018$ & $-0.00498$\\
\midrule
CIFAR-100 top-1 & 11  & 0.4942 & \best{0.5079} & 0.4841 & $+0.0137$ & $-0.0101$\\
 & 29  & \best{0.5076} & 0.5046 & 0.5016 & $-0.0030$ & $-0.0060$\\
 & 47  & \best{0.5005} & 0.4978 & 0.4937 & $-0.0027$ & $-0.0068$\\
 & 71  & 0.4908 & \best{0.4971} & 0.4940 & $+0.0063$ & $+0.0032$\\
 & 101 & 0.4985 & \best{0.4998} & 0.4885 & $+0.0013$ & $-0.0100$\\
\bottomrule
\end{tabular}

\end{table}

\begin{table}[htbp]
\centering\footnotesize
\setlength{\tabcolsep}{3pt}
\caption{Individual final official-test endpoints in the A6000 replications. ISIC16 reports mean per-image foreground Dice and CIFAR-100 reports top-1 accuracy. Each difference is F minus R; T was not included. Bold marks the higher observed score within each task and seed.}
\label{tab:replicationseeds}
\begin{tabular}{r rrr rrr}
\toprule
& \multicolumn{3}{c}{ISIC16 Dice} & \multicolumn{3}{c}{CIFAR-100 top-1}\\
\textbf{Seed} & \textbf{R} & \textbf{F} & \textbf{F--R} & \textbf{R} & \textbf{F} & \textbf{F--R}\\
\midrule
205785622 & 0.81872 & \best{0.84830} & $+0.02958$ & 0.4978 & \best{0.5064} & $+0.0086$\\
458014223 & \best{0.86188} & 0.85724 & $-0.00464$ & \best{0.5023} & 0.4879 & $-0.0144$\\
565813932 & 0.87312 & \best{0.87395} & $+0.00084$ & \best{0.4981} & 0.4970 & $-0.0011$\\
646228129 & 0.86857 & \best{0.87216} & $+0.00359$ & \best{0.5024} & 0.5007 & $-0.0017$\\
821637680 & \best{0.87377} & 0.87369 & $-0.00009$ & 0.4911 & \best{0.4959} & $+0.0048$\\
864171589 & \best{0.86830} & 0.86749 & $-0.00081$ & \best{0.4936} & 0.4902 & $-0.0034$\\
1052495617 & 0.86081 & \best{0.86092} & $+0.00011$ & \best{0.4999} & 0.4947 & $-0.0052$\\
1059645469 & 0.86592 & \best{0.87168} & $+0.00576$ & 0.4950 & \best{0.4969} & $+0.0019$\\
1147609747 & 0.85524 & \best{0.87245} & $+0.01721$ & \best{0.5058} & 0.4968 & $-0.0090$\\
1192596770 & 0.85265 & \best{0.85341} & $+0.00076$ & 0.4849 & \best{0.4881} & $+0.0032$\\
1209522243 & 0.85207 & \best{0.86467} & $+0.01260$ & 0.4992 & \best{0.5013} & $+0.0021$\\
1338995025 & \best{0.87355} & 0.86242 & $-0.01113$ & 0.4935 & \best{0.4995} & $+0.0060$\\
1619140427 & 0.85846 & \best{0.86549} & $+0.00703$ & \best{0.5074} & 0.5029 & $-0.0045$\\
1898335833 & \best{0.87013} & 0.83931 & $-0.03082$ & 0.4915 & \best{0.4928} & $+0.0013$\\
1985788232 & 0.85998 & \best{0.86942} & $+0.00944$ & \best{0.4940} & 0.4797 & $-0.0143$\\
\bottomrule
\end{tabular}
\end{table}

\FloatBarrier
\subsection{Secondary endpoints and sensitivities}

\begin{table}[htbp]
\centering\footnotesize
\caption{Secondary ISIC16 outcomes. Differences are paired candidate minus R with approximate 95\% intervals. A dash denotes an unreported mean or an inapplicable contrast. Speedup rows report R/candidate ratios in the mean column and their 95\% intervals in the final column. Bold marks the highest quality mean or largest speedup for each outcome.}
\label{tab:secondaryoutcomes}
\setlength{\tabcolsep}{4pt}
\begin{tabularx}{\linewidth}{@{}lp{0.23\linewidth}l r Y@{}}
\toprule
\textbf{GPU} & \textbf{Outcome} & \textbf{Path} & \textbf{Mean / ratio} & \textbf{Difference [95\% interval]}\\
\midrule
V100 & Duplicate-excluded Dice & F & -- & $-0.01050$ [$-0.02872$, $0.00773$]\\
 & & T & -- & $-0.00460$ [$-0.02163$, $0.01243$]\\
\addlinespace
V100 & Final foreground IoU & R & \best{0.78809} & --\\
 & & F & 0.77622 & $-0.01187$ [$-0.03710$, $0.01335$]\\
 & & T & 0.78664 & $-0.00145$ [$-0.02352$, $0.02062$]\\
\addlinespace
V100 & Validation-selected Dice & R & 0.86624 & --\\
 & & F & 0.85942 & $-0.00683$ [$-0.01380$, $0.00015$]\\
 & & T & \best{0.86871} & $+0.00246$ [$-0.00164$, $0.00657$]\\
\addlinespace
V100 & Worker elapsed speedup & F & \best{1.323} & [1.208, 1.448]\\
 & & T & 1.301 & [1.050, 1.611]\\
\midrule
A6000 & Duplicate-excluded Dice & F & -- & $+0.00292$ [$-0.00331$, $0.00915$]\\
\addlinespace
A6000 & Final foreground IoU & R & 0.78134 & --\\
 & & F & \best{0.78469} & $+0.00335$ [$-0.00485$, $0.01154$]\\
\addlinespace
A6000 & Validation-selected Dice & R & \best{0.86457} & --\\
 & & F & 0.86227 & $-0.00229$ [$-0.00684$, $0.00226$]\\
\bottomrule
\end{tabularx}
\end{table}

For V100, the additional T-minus-F contrasts are $+0.01042$ [$-0.00428$, $0.02513$] for final foreground IoU and $+0.00929$ [$0.00048$, $0.01810$] for validation-selected Dice.

\begin{table}[htbp]
\centering\footnotesize
\caption{Secondary CIFAR-100 outcomes. Differences are paired candidate minus R with approximate 95\% intervals. Cross-entropy reports mean $\pm$ SD; speedup rows report R/candidate ratios and their 95\% intervals in the final column. Bold marks the highest accuracy, lowest cross-entropy, or largest speedup for each outcome.}
\label{tab:cifarsecondary}
\setlength{\tabcolsep}{4pt}
\begin{tabularx}{\linewidth}{@{}lp{0.23\linewidth}l r Y@{}}
\toprule
\textbf{GPU} & \textbf{Outcome} & \textbf{Path} & \textbf{Mean / ratio} & \textbf{Difference [95\% interval]}\\
\midrule
V100 & Duplicate-excluded top-1 & F & -- & $+0.00146$ [$-0.00653$, $0.00946$]\\
 & & T & -- & $-0.00697$ [$-0.01356$, $-0.00037$]\\
\addlinespace
V100 & Validation-selected top-1 & R & \best{0.50244} & --\\
 & & F & 0.50236 & $-0.00008$ [$-0.01418$, $0.01402$]\\
 & & T & 0.49490 & $-0.00754$ [$-0.01908$, $0.00400$]\\
\addlinespace
V100 & Final cross-entropy & R & $1.97160\pm0.01918$ & --\\
 & & F & $\mathbf{1.96515}\pm0.02289$ & --\\
 & & T & $1.99975\pm0.02161$ & --\\
\addlinespace
V100 & Worker elapsed speedup & F & 1.100 & [0.984, 1.229]\\
 & & T & \best{1.168} & [1.045, 1.305]\\
\midrule
A6000 & Duplicate-excluded top-1 & F & -- & $-0.00105$ [$-0.00503$, $0.00292$]\\
\addlinespace
A6000 & Validation-selected top-1 & R & \best{0.49811} & --\\
 & & F & 0.49694 & $-0.00117$ [$-0.00541$, $0.00308$]\\
\bottomrule
\end{tabularx}
\end{table}

V100 R entries use the repeated runs of main-text Section~5.6; the worker elapsed speedups use the original runs. The exact sign-test value for the V100 validation-selected F-minus-R top-1 contrast is $1.0$. IoU values are taken from retained endpoint records, not converted from aggregate Dice. Validation-selected checkpoints use development data only. These secondary and duplicate-excluded results do not change the primary interpretation.

\FloatBarrier
\subsection{Scaling estimates and profile}
Across the six jointly completed scaling configurations, the smallest paired speedup is $1.294$ [1.058, 1.582] at batch 1 and $256\times256$; the largest is $2.103$ [2.073, 2.132] at batch 16 and $256\times256$. The shared batch-8, $256\times256$ speedup is $1.935$ [1.907, 1.964]. The recorded allocator limit is 25,567,444,992 bytes (23.81~GiB).

\begin{table}[htbp]

\centering\footnotesize

\caption{Descriptive 20-call batch-8 U-Net update trace. Bold marks the lowest time or launch count; reduction share is a composition measure. Cumulative kernel time sums individually recorded device events and is not bounded by module span.}

\label{tab:profile}

\begin{tabular}{lrrrr}
\toprule
\textbf{Path} & \textbf{Module span (ms)} & \textbf{Kernel time (ms/call)} & \textbf{Launches} & \textbf{Reduction share}\\
\midrule
R & 102.697 & 119.099 & 4,121 & 49.80\%\\
F & 54.369 & \best{59.060} & 3,603 & 64.59\%\\
T & \best{53.862} & 60.926 & \best{3,503} & 62.38\%\\
\bottomrule
\end{tabular}

\end{table}

\FloatBarrier
\section{Elementwise agreement and learned curvature}

Table~\ref{tab:elementwise} summarizes a separate replay of the structured suite that records the elementwise relative error $|y_{\mathrm{cand}}-y_{\mathrm{ref}}|/\max(|y_{\mathrm{ref}}|,10^{-12})$ for each compared tensor. About half of the output entries have reference magnitudes below $10^{-12}$; because no output error exceeds $3\times10^{-2}$, the absolute error at these entries is below $3\times10^{-14}$. The 32 FP32 gradient misses under F in this replay correspond to 32 of T's 33 misses in the accepted run (main-text Table~2); T's remaining miss is the separately reported curvature gradient. They come from one ReLU input whose sign is set by FP32 rounding (main-text Section~5.1); three further replays of these cases reproduced Table~2.

\begin{table}[htbp]
\centering\footnotesize
\caption{Elementwise relative errors against R in the structured replay. Passed counts use the tensor-level criterion of main-text Eq.~(6). Bold marks the highest pass count or lowest displayed error within each precision and tensor group; ties are retained. The median column is the median across tensors of each tensor's median elementwise error; the remaining columns are maxima across tensors of the per-tensor 99th and 99.9th percentiles and of the largest elementwise error.}
\label{tab:elementwise}
\setlength{\tabcolsep}{4pt}
\begin{tabular}{lllrcccc}
\toprule
\textbf{Precision} & \textbf{Path} & \textbf{Tensors} & \textbf{Passed} & \textbf{Median} & \textbf{Max p99} & \textbf{Max p99.9} & \textbf{Maximum}\\
\midrule
FP32 & F & Outputs & \best{20/20} & \best{0} & $\boldsymbol{9.3\times10^{-5}}$ & $\boldsymbol{7.6\times10^{-4}}$ & $\boldsymbol{1.8\times10^{-2}}$\\
FP32 & F & Gradients & 628/660 & $\boldsymbol{8.7\times10^{-7}}$ & $\boldsymbol{2.3\times10^{0}}$ & $\boldsymbol{3.4\times10^{0}}$ & $\boldsymbol{3.4\times10^{0}}$\\
FP32 & T & Outputs & \best{20/20} & \best{0} & $1.2\times10^{-4}$ & $1.4\times10^{-3}$ & $2.9\times10^{-2}$\\
FP32 & T & Gradients & \best{660/660} & $8.8\times10^{-7}$ & $1.3\times10^{1}$ & $3.4\times10^{1}$ & $3.8\times10^{1}$\\
\addlinespace
FP64 & F & Outputs & \best{20/20} & \best{0} & $1.8\times10^{-13}$ & $1.8\times10^{-10}$ & $3.0\times10^{-10}$\\
FP64 & F & Gradients & \best{660/660} & $\boldsymbol{1.5\times10^{-15}}$ & $1.4\times10^{-7}$ & $1.4\times10^{-7}$ & $1.4\times10^{-7}$\\
FP64 & T & Outputs & \best{20/20} & \best{0} & $\boldsymbol{1.3\times10^{-13}}$ & $\boldsymbol{1.1\times10^{-10}}$ & $\boldsymbol{1.8\times10^{-10}}$\\
FP64 & T & Gradients & \best{660/660} & $\boldsymbol{1.5\times10^{-15}}$ & $\boldsymbol{9.0\times10^{-8}}$ & $\boldsymbol{1.0\times10^{-7}}$ & $\boldsymbol{1.2\times10^{-7}}$\\
\bottomrule
\end{tabular}
\end{table}

\begin{figure}[htbp]
\centering
\includegraphics[width=\linewidth]{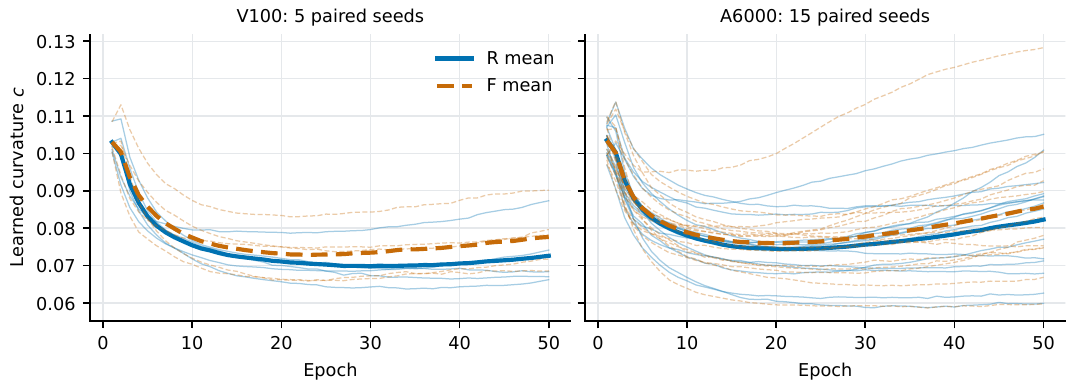}
\caption{Learned ISIC16 curvature at the end of each epoch. Thin lines show individual runs and thick lines the mean over seeds; R is solid and F dashed.}
\label{fig:curvature}
\end{figure}

Table~\ref{tab:curvature} lists the final learned curvature of every ISIC16 seed pair, and Figure~\ref{fig:curvature} shows the trajectories. All runs start from $c=0.1$.

\begin{table}[htbp]
\centering\footnotesize
\caption{Learned ISIC16 curvature $c$ at the fixed final update for each seed pair. The mean F-minus-R difference is $+0.0050$ [$-0.0104$, $0.0205$] for V100 and $+0.0035$ [$-0.0020$, $0.0089$] for A6000 (paired 95\% intervals).}
\label{tab:curvature}
\begin{tabular}{llrrr}
\toprule
\textbf{GPU} & \textbf{Seed} & \textbf{R} & \textbf{F} & \textbf{F--R}\\
\midrule
V100 & 11 & 0.0670 & 0.0686 & $+0.0016$\\
 & 29 & 0.0684 & 0.0901 & $+0.0218$\\
 & 47 & 0.0873 & 0.0782 & $-0.0092$\\
 & 71 & 0.0741 & 0.0717 & $-0.0024$\\
 & 101 & 0.0662 & 0.0796 & $+0.0134$\\
\midrule
A6000 & 205785622 & 0.0892 & 0.0829 & $-0.0063$\\
 & 458014223 & 0.0886 & 0.1006 & $+0.0120$\\
 & 565813932 & 0.0770 & 0.0749 & $-0.0022$\\
 & 646228129 & 0.1009 & 0.0927 & $-0.0082$\\
 & 821637680 & 0.0626 & 0.0747 & $+0.0121$\\
 & 864171589 & 0.1051 & 0.1283 & $+0.0232$\\
 & 1052495617 & 0.0712 & 0.0781 & $+0.0068$\\
 & 1059645469 & 0.0884 & 0.0863 & $-0.0021$\\
 & 1147609747 & 0.0599 & 0.0597 & $-0.0001$\\
 & 1192596770 & 0.0718 & 0.0801 & $+0.0083$\\
 & 1209522243 & 0.0851 & 0.1003 & $+0.0152$\\
 & 1338995025 & 0.0922 & 0.0898 & $-0.0024$\\
 & 1619140427 & 0.0679 & 0.0669 & $-0.0011$\\
 & 1898335833 & 0.0875 & 0.0752 & $-0.0123$\\
 & 1985788232 & 0.0867 & 0.0959 & $+0.0092$\\
\bottomrule
\end{tabular}
\end{table}

Table~\ref{tab:oracle} compares the FP32 errors of R, F, and T with an FP64 oracle, R evaluated in FP64 with the FP32 projection margin, on the 20 structured realizations. The tested patches have normalized radii of at most 0.75, so the radius cap is inactive. The large maxima occur in all paths for tiny, cancellation-dominated tensors such as curvature gradients and scalar normalization weights.

\begin{table}[htbp]
\centering\footnotesize
\caption{FP32 error against the FP64 oracle at $c=0.1$, computed on CPU. The final column gives the median relative L2 gradient error at $c=0.06$ and $c=0.13$, with a shared factor of $10^{-6}$. Bold marks the lowest displayed error within each tensor group and metric; ties are retained.}
\label{tab:oracle}
\setlength{\tabcolsep}{3pt}
\begin{tabular}{llccccc}
\toprule
\textbf{Tensors} & \textbf{Path} & \textbf{Median rel.\ L2} & \textbf{Max rel.\ L2} & \textbf{Max abs.} & \textbf{Median p99} & \textbf{$c=0.06$\,/\,$0.13$}\\
\midrule
Outputs (20) & R & $1.13\times10^{-6}$ & $\boldsymbol{1.69\times10^{-6}}$ & $3.9\times10^{-6}$ & $2.6\times10^{-5}$ & --\\
& F & $1.10\times10^{-6}$ & $2.01\times10^{-6}$ & $3.7\times10^{-6}$ & $\boldsymbol{2.5\times10^{-5}}$ & --\\
& T & $\boldsymbol{0.98\times10^{-6}}$ & $2.04\times10^{-6}$ & $\boldsymbol{3.4\times10^{-6}}$ & $2.7\times10^{-5}$ & --\\
\addlinespace
Gradients (660) & R & $1.05\times10^{-6}$ & $1.49\times10^{-2}$ & $2.7\times10^{-4}$ & $7.3\times10^{-6}$ & $(1.15\,/\,1.06)\times10^{-6}$\\
& F & $1.05\times10^{-6}$ & $\boldsymbol{0.98\times10^{-2}}$ & $3.0\times10^{-4}$ & $7.6\times10^{-6}$ & $(1.08\,/\,1.03)\times10^{-6}$\\
& T & $\boldsymbol{1.00\times10^{-6}}$ & $3.25\times10^{-2}$ & $\boldsymbol{2.4\times10^{-4}}$ & $\boldsymbol{7.2\times10^{-6}}$ & $\boldsymbol{(1.04\,/\,0.96)\times10^{-6}}$\\
\bottomrule
\end{tabular}
\end{table}

\FloatBarrier
\section{Official HypLL implementation}

Table~\ref{tab:hypll} compares the official \texttt{HConvolution2d} layer of HypLL 0.1.1 with a PyTorch implementation of F that uses the same parameters, on one RTX A6000 (PyTorch 2.4.0, CUDA 12.1, FP32, TF32 disabled). Each workload uses five fresh-process sessions per path in randomized order, with 50 warmup and 200 synchronized calls; speedups are geometric means of paired session ratios with 95\% intervals. Peak allocation is the steady-state allocator peak after warmup, which equaled the process peak within 0.25~MiB. Peak-allocation ratios computed from unrounded values are 3.87, 3.87, and 8.99 for the forward-plus-gradient calls at S1--S3 and 1.99 for the training step. Gradient calls differentiate the mean squared output. The complete-model rows use the official Hyperbolic U-Net (batch 8, $256\times256$, depth 4, widths 8--128) with all 19 convolutions replaced; its training step uses the Dice and focal losses, gradient clipping at norm 1, and Riemannian Adam.

\begin{table}[htbp]
\centering\footnotesize
\caption{Official HypLL convolution versus factorized PyTorch implementation. Times are medians of session medians. Bold marks the lower latency and allocation within each row and speedup estimates whose intervals are entirely above one. Allocation pairs are HypLL\,/\,F.}
\label{tab:hypll}
\begin{tabular}{llrrcr}
\toprule
\textbf{Workload} & \textbf{Boundary} & \textbf{HypLL (ms)} & \textbf{F (ms)} & \textbf{Speedup [95\% interval]} & \textbf{Peak (GiB)}\\
\midrule
S1 & forward & 0.988 & \best{0.645} & \best{1.524} [1.493, 1.555] & 0.084\,/\,\best{0.020}\\
 & forward + gradient & 6.693 & \best{3.976} & \best{1.426} [1.042, 1.953] & 0.184\,/\,\best{0.048}\\
S2 & forward & 25.084 & \best{5.077} & \best{4.940} [4.932, 4.947] & 2.387\,/\,\best{0.575}\\
 & forward + gradient & 152.205 & \best{24.325} & \best{6.255} [6.247, 6.264] & 5.288\,/\,\best{1.368}\\
S3 & forward & 69.746 & \best{7.301} & \best{9.555} [9.549, 9.562] & 7.137\,/\,\best{0.825}\\
 & forward + gradient & 422.852 & \best{36.232} & \best{11.673} [11.663, 11.684] & 15.663\,/\,\best{1.743}\\
\addlinespace
Hyperbolic U-Net & forward & 114.0 & \best{82.9} & \best{1.309} [1.145, 1.498] & 1.28\,/\,\best{0.28}\\
 & training step & 547.2 & \best{386.6} & \best{1.413} [1.409, 1.418] & 13.61\,/\,\best{6.83}\\
\bottomrule
\end{tabular}
\end{table}

With identical parameters, the factorized layer meets the agreement criterion in all 80 output and gradient comparisons on S1--S3 and the stride-2 check: the relative L2 error is at most $1.0\times10^{-14}$ in FP64 and $4.3\times10^{-6}$ in FP32. Separate stress tests, documented in the code release, expose shared curvature-gradient error near the ball boundary and output-map overflow and are excluded from these agreement bounds. In one forward session of the official U-Net, its iterative batch-normalization mean required about 1,000 iterations per pass in both paths, which slowed both equally and widens that interval.

\FloatBarrier
\section{Cap activity and retained-state controls}
Table~\ref{tab:capactivity} gives the scope of the cap-activity statement in main-text Section~3. Table~\ref{tab:retainedstate} gives the retained-state gradient differences and controls cited in main-text Section~5.1.

\begin{table}[htbp]
\centering\small
\caption{Source-radius cap activity in trained models. Each row covers the final checkpoints (update 5,100 for ISIC16 and 35,200 for CIFAR-100) of the five V100 training seeds 11, 29, 47, 71, and 101 of one path. Each checkpoint is evaluated in its own path on the complete official test set in manifest order with the registered evaluation batch size (ISIC16: 379 images, batch 8; CIFAR-100: 10,000 images, batch 128), in FP32 on CPU. At every output position of every Poincar\'e convolution, the normalized source radius $\rho=\tanh(\sqrt{c}\,\norm{P})$ of the scaled tangent patch $P$ is computed with F's expression from the layer's actual input; the cap is active when $\rho\geq 1-0.004$. Fractions count cap-active output positions. The all-layer fraction pools all layers, checkpoints, and test images. The largest layer fraction is pooled over the five checkpoints or taken from one checkpoint. The first convolution is the U-Net input convolution and the ResNet-20 stem. All cap-active positions occur in one ResNet-20 layer, the first convolution of the second block in the last stage, in two checkpoints per path. T was not evaluated.}
\label{tab:capactivity}
\setlength{\tabcolsep}{3pt}
\begin{tabular}{@{}llccccc@{}}
\toprule
 & & & \multicolumn{2}{c}{\textbf{Largest layer fraction}} & \multicolumn{2}{c}{\textbf{Largest $\rho$}}\\
\cmidrule(lr){4-5}\cmidrule(l){6-7}
\textbf{Task} & \textbf{Path} & \textbf{All-layer fraction} & \textbf{Pooled} & \textbf{One checkpoint} & \textbf{All layers} & \textbf{First conv.}\\
\midrule
ISIC16 U-Net & R & 0 & 0 & 0 & 0.9425 & 0.4430\\
 & F & 0 & 0 & 0 & 0.9224 & 0.4491\\
\addlinespace
CIFAR-100 ResNet-20 & R & $1.3\times10^{-8}$ & $1.9\times10^{-6}$ & $6.3\times10^{-6}$ & 0.9964 & 0.4694\\
 & F & $2.6\times10^{-8}$ & $3.8\times10^{-6}$ & $1.7\times10^{-5}$ & 0.9964 & 0.4694\\
\bottomrule
\end{tabular}
\end{table}

\begin{table}[htbp]
\centering\small
\caption{Loss-gradient differences from R at retained training states. Rows give the seed-11 training run whose final state was evaluated. Entries are relative L2 differences from R over the complete raw loss-gradient tree before clipping, L2 regularization, and Adam (486,941 U-Net and 71,727 ResNet-20 entries), computed in FP32 on one V100 for the same first post-budget training batch. R repeat evaluates R's compiled gradient function a second time on the identical state and batch. R one-ulp evaluates R after every floating-point parameter element is moved by one unit in the last place, alternately toward $+\infty$ and $-\infty$, with the optimizer state unchanged; it is a declared sensitivity control, not a model of hardware nondeterminism. A zero entry is bitwise identical to R. The ResNet-20 differences, including the nonzero repeated-R entries, come from the miscompiled stem-weight gradient of R described in main-text Section~5.1; excluding that tensor, the F--R differences are 0.05--0.19\%.}
\label{tab:retainedstate}
\setlength{\tabcolsep}{4pt}
\begin{tabular}{@{}llcccc@{}}
\toprule
 & & \multicolumn{2}{c}{\textbf{Paths}} & \multicolumn{2}{c}{\textbf{R controls}}\\
\cmidrule(lr){3-4}\cmidrule(l){5-6}
\textbf{Task} & \textbf{State} & \textbf{F} & \textbf{T} & \textbf{R repeat} & \textbf{R one-ulp}\\
\midrule
ISIC16 U-Net & R & $7.1\times10^{-5}$ & $5.0\times10^{-6}$ & 0 & $4.2\times10^{-4}$\\
 & F & $3.2\times10^{-6}$ & $1.6\times10^{-5}$ & 0 & $1.2\times10^{-5}$\\
 & T & $5.8\times10^{-6}$ & $1.0\times10^{-4}$ & 0 & $3.7\times10^{-4}$\\
\addlinespace
CIFAR-100 ResNet-20 & R & $5.8\times10^{-2}$ & $5.8\times10^{-2}$ & $1.0\times10^{-2}$ & $1.1\times10^{-2}$\\
 & F & $6.7\times10^{-2}$ & $6.7\times10^{-2}$ & $5.9\times10^{-3}$ & $1.7\times10^{-2}$\\
 & T & $1.1\times10^{-1}$ & $1.1\times10^{-1}$ & $1.0\times10^{-2}$ & $1.2\times10^{-2}$\\
\bottomrule
\end{tabular}
\end{table}

\FloatBarrier
\section{Gradients at trained states}
Table~\ref{tab:trainedoracle} compares FP32 gradients with an FP64 reference at 40 retained checkpoints of the V100 training runs, each evaluated on its first training batch after the checkpoint. The reference is F in FP64. On the GPU it carries a relative floor of about $2\times10^{-8}$ because one scale constant is rounded to FP32 inside the native convolution; at an inspected state it agrees with R in FP64 on the CPU to this level and with central finite differences of the FP64 loss. Table~\ref{tab:stemnorms} reports the stem-weight norms that reveal the effect of R's miscompiled stem gradient on training.

\begin{table}[htbp]
\centering\footnotesize
\caption{Relative L2 error of FP32 gradients against F in FP64 at retained training states: median (maximum) over states. ResNet-20 states are produced by R, F, and T for five seeds at mid-training and at the final update; U-Net states are produced by R and F at the final update. ``R, no stem'' omits the stem-convolution weight gradient. The last column compares R in FP64 with the reference.}
\label{tab:trainedoracle}
\setlength{\tabcolsep}{4pt}
\begin{tabular}{@{}lccccc@{}}
\toprule
\textbf{States} & \textbf{F} & \textbf{T} & \textbf{R} & \textbf{R, no stem} & \textbf{R in FP64}\\
\midrule
ResNet-20, final (15) & 8.7e-4 (3.0e-3) & 5.5e-4 (1.9e-3) & 0.11 (0.16) & 6.5e-4 (3.2e-3) & 0.042 (0.11)\\
ResNet-20, mid (15) & 3.1e-4 (7.1e-4) & 3.7e-4 (1.9e-3) & 0.070 (0.13) & 3.7e-4 (1.4e-3) & 0.024 (0.050)\\
U-Net, final (10) & 3.0e-5 (4.2e-4) & 3.3e-5 (7.8e-4) & 2.0e-5 (1.3e-4) & -- & 6.6e-15 (1.2e-14)\\
\bottomrule
\end{tabular}
\end{table}

\begin{table}[htbp]
\centering\small
\caption{Largest column norm of the stem-convolution weight, median [minimum, maximum] across seeds. Initial values are identical across paths. Only the V100 CIFAR-100 R runs, which used R's miscompiled stem gradient, grow beyond the other runs; their growth occurs between the mid-training and final checkpoints. The largest column norm of a late ResNet-20 convolution is 2.5--3.1 at the final checkpoint of every run.}
\label{tab:stemnorms}
\begin{tabular}{@{}llccc@{}}
\toprule
\textbf{Cohort} & \textbf{Checkpoint} & \textbf{R} & \textbf{F} & \textbf{T}\\
\midrule
CIFAR-100, V100 (5 seeds) & Mid-training & 1.00 [0.91, 1.14] & 0.79 [0.74, 0.87] & 0.81 [0.78, 0.98]\\
 & Final & 21.2 [6.17, 61.3] & 1.08 [1.03, 1.22] & 1.10 [1.03, 1.30]\\
 & Final, repeated R & 1.16 [1.02, 1.51] & -- & --\\
CIFAR-100, A6000 (15 seeds) & Final & 1.18 [1.05, 1.49] & 1.10 [0.96, 1.63] & --\\
\addlinespace
ISIC16, V100 (5 seeds) & Final & 0.46 [0.45, 0.50] & 0.48 [0.47, 0.52] & 0.48 [0.48, 0.49]\\
ISIC16, A6000 (15 seeds) & Final & 0.47 [0.43, 0.53] & 0.49 [0.43, 0.58] & --\\
\bottomrule
\end{tabular}
\end{table}

\FloatBarrier
\section{Explicit references under JAX 0.11.2}
Table~\ref{tab:jax0112} lists the JAX 0.11.2 medians behind main-text Table~6. The measurements use one RTX A6000 (CUDA-12 plugin, cuDNN 9.27.0, allocator fraction 0.95), five fresh-process paired sessions per group, and the protocol of the JAX 0.4.25 controls. During timing, the only other process on the device, a training job of the authors holding 16.6~GiB, was suspended. Table~\ref{tab:jax0112trace} summarizes 20-call device traces of the batch-8 U-Net update for Rg and F under both JAX versions on the same device. Under JAX 0.11.2, Rg's kernel time decreases but each call waits a median of 24~ms before its first kernel, against 0.8~ms under JAX 0.4.25. Kernel categories are assigned by name; under JAX 0.11.2 most kernels run inside CUDA-graph command buffers.

\begin{table}[htbp]
\centering\small
\caption{Median latency (ms) and peak live allocation (GiB) under JAX 0.11.2. R, Rg, Mg, and F share paired sessions; E comes from separate paired R/F/E sessions, in which R and F times differ from those shown by at most 0.5\%.}
\label{tab:jax0112}
\setlength{\tabcolsep}{4pt}
\begin{tabular}{@{}lccccc@{}}
\toprule
\textbf{Workload} & \textbf{R} & \textbf{Rg} & \textbf{Mg} & \textbf{F} & \textbf{E}\\
\midrule
U-Net, batch 8, forward & 33.77 / 0.963 & 23.97 / 0.596 & 17.59 / 0.686 & 19.81 / 0.619 & 3.48 / 0.619\\
U-Net, batch 8, update & 114.74 / 4.687 & 104.57 / 4.352 & 85.81 / 3.443 & 68.87 / 2.386 & 11.96 / 0.639\\
ResNet-20, batch 128, update & 35.40 / 1.537 & 43.94 / 1.485 & 29.48 / 1.150 & 20.74 / 0.817 & 4.32 / 0.221\\
\bottomrule
\end{tabular}
\end{table}

\begin{table}[htbp]
\centering\small
\caption{Cumulative kernel time per batch-8 U-Net update (ms) by category, under JAX 0.4.25 and 0.11.2 on the same RTX A6000. ``Other'' includes transcendental kernels. Categories describe trace attribution, not exclusive hardware occupancy.}
\label{tab:jax0112trace}
\begin{tabular}{@{}lcccc@{}}
\toprule
 & \multicolumn{2}{c}{\textbf{Rg}} & \multicolumn{2}{c}{\textbf{F}}\\
\cmidrule(lr){2-3}\cmidrule(l){4-5}
\textbf{Category} & \textbf{0.4.25} & \textbf{0.11.2} & \textbf{0.4.25} & \textbf{0.11.2}\\
\midrule
Convolution & -- & -- & 5.5 & 5.3\\
Matrix multiplication & 10.0 & 6.5 & 0.3 & 0.3\\
Reduction & 51.2 & 67.8 & 32.1 & 48.1\\
Transpose & -- & -- & 2.3 & 2.5\\
Other & 26.8 & 5.3 & 6.8 & 9.5\\
All kernels & 87.9 & 79.7 & 46.9 & 65.7\\
\bottomrule
\end{tabular}
\end{table}

\FloatBarrier
\section{Identity of the gather-based reference}
Table~\ref{tab:rgidentity} compares Rg with R at the seed-11 initialization and at the retained final states of the V100 training runs, on real training batches and with the pinned JAX 0.4.25 stack on V100. A repeated evaluation of R serves as a control.

\begin{table}[htbp]
\centering\small
\caption{Pooled relative L2 differences from R of compiled Rg and of a repeated R evaluation, at initialization / final state. Patch tensors of all convolutions are bitwise identical between R and Rg in both models and states. ResNet-20 entries separate the stem-convolution weight gradient, which R computes incorrectly and nondeterministically on this stack (main-text Section~5.1). Parameters are compared after 20 updates with identical data order.}
\label{tab:rgidentity}
\setlength{\tabcolsep}{4pt}
\begin{tabular}{@{}llcc@{}}
\toprule
\textbf{Model} & \textbf{Quantity} & \textbf{Rg} & \textbf{R repeated}\\
\midrule
U-Net & Gradients & $1.2\times10^{-4}$ / $5.1\times10^{-6}$ & 0 / 0\\
 & Parameters after 20 updates & 0.046 / $3.5\times10^{-5}$ & 0 / 0\\
\addlinespace
ResNet-20 & Stem weight gradient & 0.027 / 0.27 & $1.9\times10^{-3}$ / 0.032\\
 & Other gradients & 0.033 / $2.0\times10^{-4}$ & 0 / 0\\
 & Parameters after 20 updates & 0.089 / $1.1\times10^{-3}$ & $6.9\times10^{-3}$ / $2.8\times10^{-4}$\\
\bottomrule
\end{tabular}
\end{table}